\documentclass[10pt,twocolumn,letterpaper]{article}
\usepackage{cvpr}
\usepackage{times}
\usepackage{epsfig}
\usepackage{graphicx}
\usepackage{gensymb} 
\usepackage{color}
\usepackage[pagebackref=true,breaklinks=true,letterpaper=true,colorlinks,bookmarks=false]{hyperref}
\usepackage{amsmath,amsfonts,amssymb,mathrsfs}

\usepackage{graphicx}
\usepackage{amsmath,amssymb}
\usepackage{epstopdf}
\usepackage{tabularx}
\usepackage{eqnarray}
\usepackage{array,booktabs}
\usepackage{enumitem}
\usepackage[space]{grffile} 
\usepackage{soul} 
\usepackage{float}
\usepackage{tabularx,ragged2e} 
\usepackage{bbm}

\definecolor{myGray}{rgb}{0.6,0.6,0.6}

\newcommand{\symbQ}{\mathsf{Q}}
\newcommand{\symbS}{\mathsf{S}}

\newcommand{\sheq}{=}
\newcommand{\shg}{\hspace{-0.4ex}>\hspace{-0.4ex}}

\newcommand{\norm}[1]{\lVert#1\rVert}
\DeclareMathOperator*{\gru}{GRU}
\DeclareMathOperator*{\pool}{pool}

\DeclareMathOperator*{\argmax}{arg\,max}

\newcommand{\nospace}{\hspace{-0.35em}}

\newcommand{\red}[1]{\textcolor{red}{#1}}

\def\onedot{\ifx\@let@token.\else.\null\fi\xspace}

\def\eg{\emph{e.g}\onedot} 
\def\ie{\emph{i.e}\onedot} 
 
\def\etc{\emph{etc}\onedot} 
\def\etal{\emph{et al}\onedot}

\newcommand{\fig}{Fig.~}
\newcommand{\eq}{Eq.\,}
\newcommand{\sect}{Section~}
\newcommand{\tab}{Table~}

\newcommand{\hangBox}[1]{
  \begin{minipage}[t]{\textwidth}
  \begin{tabbing} 
  ~\\[-\baselineskip] 
  #1 
  \end{tabbing}
  \end{minipage}}

\newcommand{\resultsAttentionAbstractPage}[8]{
  \begin{figure}[t]
  \renewcommand{\tabcolsep}{0.5mm}
  \renewcommand{\arraystretch}{1.0}
  \begin{tabularx}{\textwidth}{*{4}{>{\raggedright\arraybackslash}X}}
  
  \hangBox{
    \includegraphics[width=42mm]{#1.jpg}\\
    \includegraphics[width=42mm]{#1-2.jpg}\\
  }&
    \begin{minipage}{\linewidth}\footnotesize\input{"#1.txt"}\end{minipage}
    \vspace{7mm}
    \hangBox{\includegraphics[width=40mm]{#1-attention.pdf}}
  &
  \hangBox{
    \includegraphics[width=42mm]{#2.jpg}\\
    \includegraphics[width=42mm]{#2-2.jpg}\\
  }&
    \begin{minipage}{\linewidth}\footnotesize\input{"#2.txt"}\end{minipage}
    \vspace{7mm}
    \hangBox{\includegraphics[width=40mm]{#2-attention.pdf}}
  \\
\\
  
  \hangBox{
    \includegraphics[width=42mm]{#3.jpg}\\
    \includegraphics[width=42mm]{#3-2.jpg}\\
  }&
    \begin{minipage}{\linewidth}\footnotesize\input{"#3.txt"}\end{minipage}
    \vspace{7mm}
    \hangBox{\includegraphics[width=40mm]{#3-attention.pdf}}
  &
  \hangBox{
    \includegraphics[width=42mm]{#4.jpg}\\
    \includegraphics[width=42mm]{#4-2.jpg}\\
  }&
    \begin{minipage}{\linewidth}\footnotesize\input{"#4.txt"}\end{minipage}
    \vspace{7mm}
    \hangBox{\includegraphics[width=40mm]{#4-attention.pdf}}
  \\
\\
  
  \hangBox{
    \includegraphics[width=42mm]{#5.jpg}\\
    \includegraphics[width=42mm]{#5-2.jpg}\\
  }&
    \begin{minipage}{\linewidth}\footnotesize\input{"#5.txt"}\end{minipage}
    \vspace{7mm}
    \hangBox{\includegraphics[width=40mm]{#5-attention.pdf}}
  &
  \hangBox{
    \includegraphics[width=42mm]{#6.jpg}\\
    \includegraphics[width=42mm]{#6-2.jpg}\\
  }&
    \begin{minipage}{\linewidth}\footnotesize\input{"#6.txt"}\end{minipage}
    \vspace{7mm}
    \hangBox{\includegraphics[width=40mm]{#6-attention.pdf}}
  \\
\\
  
  \hangBox{
    \includegraphics[width=42mm]{#7.jpg}\\
    \includegraphics[width=42mm]{#7-2.jpg}\\
  }&
    \begin{minipage}{\linewidth}\footnotesize\input{"#7.txt"}\end{minipage}
    \vspace{7mm}
    \hangBox{\includegraphics[width=40mm]{#7-attention.pdf}}
  &
  \hangBox{
    \includegraphics[width=42mm]{#8.jpg}\\
    \includegraphics[width=42mm]{#8-2.jpg}\\
  }&
    \begin{minipage}{\linewidth}\footnotesize\input{"#8.txt"}\end{minipage}
    \vspace{7mm}
    \hangBox{\includegraphics[width=40mm]{#8-attention.pdf}}
  \\

  \end{tabularx}
  \end{figure}
}

\newcommand{\resultsAttentionAbstractPageSix}[6]{
  \begin{figure}[H]
  \renewcommand{\tabcolsep}{0.5mm}
  \renewcommand{\arraystretch}{1.0}
  \begin{tabularx}{\textwidth}{*{4}{>{\raggedright\arraybackslash}X}}
  
  \hangBox{
    \includegraphics[width=42mm]{#1.jpg}\\
    \includegraphics[width=42mm]{#1-2.jpg}\\
  }&
    \begin{minipage}{\linewidth}\footnotesize\input{"#1.txt"}\end{minipage}
    \vspace{7mm}
    \hangBox{\includegraphics[width=40mm]{#1-attention.pdf}}
  &
  \hangBox{
    \includegraphics[width=42mm]{#2.jpg}\\
    \includegraphics[width=42mm]{#2-2.jpg}\\
  }&
    \begin{minipage}{\linewidth}\footnotesize\input{"#2.txt"}\end{minipage}
    \vspace{7mm}
    \hangBox{\includegraphics[width=40mm]{#2-attention.pdf}}
  \\
\\
  
  \hangBox{
    \includegraphics[width=42mm]{#3.jpg}\\
    \includegraphics[width=42mm]{#3-2.jpg}\\
  }&
    \begin{minipage}{\linewidth}\footnotesize\input{"#3.txt"}\end{minipage}
    \vspace{7mm}
    \hangBox{\includegraphics[width=40mm]{#3-attention.pdf}}
  &
  \hangBox{
    \includegraphics[width=42mm]{#4.jpg}\\
    \includegraphics[width=42mm]{#4-2.jpg}\\
  }&
    \begin{minipage}{\linewidth}\footnotesize\input{"#4.txt"}\end{minipage}
    \vspace{7mm}
    \hangBox{\includegraphics[width=40mm]{#4-attention.pdf}}
  \\
\\
  
  \hangBox{
    \includegraphics[width=42mm]{#5.jpg}\\
    \includegraphics[width=42mm]{#5-2.jpg}\\
  }&
    \begin{minipage}{\linewidth}\footnotesize\input{"#5.txt"}\end{minipage}
    \vspace{7mm}
    \hangBox{\includegraphics[width=40mm]{#5-attention.pdf}}
  &
  \hangBox{
    \includegraphics[width=42mm]{#6.jpg}\\
    \includegraphics[width=42mm]{#6-2.jpg}\\
  }&
    \begin{minipage}{\linewidth}\footnotesize\input{"#6.txt"}\end{minipage}
    \vspace{7mm}
    \hangBox{\includegraphics[width=40mm]{#6-attention.pdf}}
  \\

  \end{tabularx}
  \end{figure}
}

\newcommand{\resultsAttentionBalancedPage}[8]{
  \begin{figure}[t]
  \renewcommand{\tabcolsep}{0.5mm}
  \renewcommand{\arraystretch}{1.0}
  \begin{tabularx}{\textwidth}{*{4}{>{\raggedright\arraybackslash}X}}
  
  \hangBox{
    \includegraphics[width=42mm]{#1.jpg}\\
    \includegraphics[width=42mm]{#1-2.jpg}\\
  }&
    \begin{minipage}{\linewidth}\footnotesize\input{"#1.txt"}\end{minipage}
    \vspace{7mm}
    \hangBox{\includegraphics[width=40mm]{#1-attention.pdf}}
  &
  \hangBox{
    \includegraphics[width=42mm]{#2.jpg}\\
    \includegraphics[width=42mm]{#2-2.jpg}\\
  }&
    \begin{minipage}{\linewidth}\footnotesize\input{"#2.txt"}\end{minipage}
    \vspace{7mm}
    \hangBox{\includegraphics[width=40mm]{#2-attention.pdf}}
\\
  
  \hangBox{
    \includegraphics[width=42mm]{#3.jpg}\\
    \includegraphics[width=42mm]{#3-2.jpg}\\
  }&
    \begin{minipage}{\linewidth}\footnotesize\input{"#3.txt"}\end{minipage}
    \vspace{7mm}
    \hangBox{\includegraphics[width=40mm]{#3-attention.pdf}}
  &
  \hangBox{
    \includegraphics[width=42mm]{#4.jpg}\\
    \includegraphics[width=42mm]{#4-2.jpg}\\
  }&
    \begin{minipage}{\linewidth}\footnotesize\input{"#4.txt"}\end{minipage}
    \vspace{7mm}
    \hangBox{\includegraphics[width=40mm]{#4-attention.pdf}}
\\
  
  \hangBox{
    \includegraphics[width=42mm]{#5.jpg}\\
    \includegraphics[width=42mm]{#5-2.jpg}\\
  }&
    \begin{minipage}{\linewidth}\footnotesize\input{"#5.txt"}\end{minipage}
    \vspace{7mm}
    \hangBox{\includegraphics[width=40mm]{#5-attention.pdf}}
  &
  \hangBox{
    \includegraphics[width=42mm]{#6.jpg}\\
    \includegraphics[width=42mm]{#6-2.jpg}\\
  }&
    \begin{minipage}{\linewidth}\footnotesize\input{"#6.txt"}\end{minipage}
    \vspace{7mm}
    \hangBox{\includegraphics[width=40mm]{#6-attention.pdf}}
\\
  
  \hangBox{
    \includegraphics[width=42mm]{#7.jpg}\\
    \includegraphics[width=42mm]{#7-2.jpg}\\
  }&
    \begin{minipage}{\linewidth}\footnotesize\input{"#7.txt"}\end{minipage}
    \vspace{7mm}
    \hangBox{\includegraphics[width=40mm]{#7-attention.pdf}}
  &
  \hangBox{
    \includegraphics[width=42mm]{#8.jpg}\\
    \includegraphics[width=42mm]{#8-2.jpg}\\
  }&
    \begin{minipage}{\linewidth}\footnotesize\input{"#8.txt"}\end{minipage}
    \vspace{7mm}
    \hangBox{\includegraphics[width=40mm]{#8-attention.pdf}}

  \end{tabularx}
  \end{figure}
}

\newcommand{\resultsAttentionBalancedPageSix}[6]{
  \begin{figure}[H]
  \renewcommand{\tabcolsep}{0.5mm}
  \renewcommand{\arraystretch}{1.0}
  \begin{tabularx}{\textwidth}{*{4}{>{\raggedright\arraybackslash}X}}
  
  \hangBox{
    \includegraphics[width=42mm]{#1.jpg}\\
    \includegraphics[width=42mm]{#1-2.jpg}\\
  }&
    \begin{minipage}{\linewidth}\footnotesize\input{"#1.txt"}\end{minipage}
    \vspace{7mm}
    \hangBox{\includegraphics[width=40mm]{#1-attention.pdf}}
  &
  \hangBox{
    \includegraphics[width=42mm]{#2.jpg}\\
    \includegraphics[width=42mm]{#2-2.jpg}\\
  }&
    \begin{minipage}{\linewidth}\footnotesize\input{"#2.txt"}\end{minipage}
    \vspace{7mm}
    \hangBox{\includegraphics[width=40mm]{#2-attention.pdf}}
\\
  
  \hangBox{
    \includegraphics[width=42mm]{#3.jpg}\\
    \includegraphics[width=42mm]{#3-2.jpg}\\
  }&
    \begin{minipage}{\linewidth}\footnotesize\input{"#3.txt"}\end{minipage}
    \vspace{7mm}
    \hangBox{\includegraphics[width=40mm]{#3-attention.pdf}}
  &
  \hangBox{
    \includegraphics[width=42mm]{#4.jpg}\\
    \includegraphics[width=42mm]{#4-2.jpg}\\
  }&
    \begin{minipage}{\linewidth}\footnotesize\input{"#4.txt"}\end{minipage}
    \vspace{7mm}
    \hangBox{\includegraphics[width=40mm]{#4-attention.pdf}}
\\
  
  \hangBox{
    \includegraphics[width=42mm]{#5.jpg}\\
    \includegraphics[width=42mm]{#5-2.jpg}\\
  }&
    \begin{minipage}{\linewidth}\footnotesize\input{"#5.txt"}\end{minipage}
    \vspace{7mm}
    \hangBox{\includegraphics[width=40mm]{#5-attention.pdf}}
  &
  \hangBox{
    \includegraphics[width=42mm]{#6.jpg}\\
    \includegraphics[width=42mm]{#6-2.jpg}\\
  }&
    \begin{minipage}{\linewidth}\footnotesize\input{"#6.txt"}\end{minipage}
    \vspace{7mm}
    \hangBox{\includegraphics[width=40mm]{#6-attention.pdf}}

  \end{tabularx}
  \end{figure}
}

\cvprfinalcopy
\def\cvprPaperID{123}

\ifcvprfinal\fi

\begin{document}

\title{Graph-Structured Representations for Visual Question Answering}

\author{Damien Teney~~~~~~Lingqiao Liu~~~~~~Anton van den Hengel\\
Australian Centre for Visual Technologies\\
The University of Adelaide\\
{\tt\small \{damien.teney,lingqiao.liu,anton.vandenhengel\}@adelaide.edu.au}
}

\maketitle


\begin{abstract}
This paper proposes to improve visual question answering (VQA) with structured representations of both scene contents and questions. A key challenge in VQA is to require joint reasoning over the visual and text domains. The predominant CNN/LSTM-based approach to VQA is limited by monolithic vector representations that largely ignore structure in the scene and in the question. CNN feature vectors cannot effectively capture situations as simple as multiple object instances, and LSTMs process questions as series of words, which do not reflect the true complexity of language structure. We instead propose to build graphs over the scene objects and over the question words, and we describe a deep neural network that exploits the structure in these representations. We show that this approach achieves significant improvements over the state-of-the-art, increasing accuracy from 71.2\% to 74.4\% on the ``abstract scenes'' multiple-choice benchmark, and from 34.7\% to 39.1\% for the more challenging ``balanced'' scenes, \ie image pairs with fine-grained differences and opposite yes/no answers to a same question.
\end{abstract}


\section{Introduction}

The task of Visual Question Answering has received growing interest in the recent years (see \cite{malinowski2014multi,antol2015vqa,wu2016survey} for example). One of the more interesting aspects of the problem is that it combines computer vision, natural language processing, and artificial intelligence. In its \textit{open-ended} form, a question is provided as text in natural language together with an image, and a correct answer must be predicted, typically in the form of a single word or a short phrase. In the \textit{multiple-choice} variant, an answer is selected from a provided set of candidates, alleviating evaluation issues related to synonyms and paraphrasing. 

\begin{figure}[t]
  \begin{center}
  \includegraphics[width=0.99\linewidth]{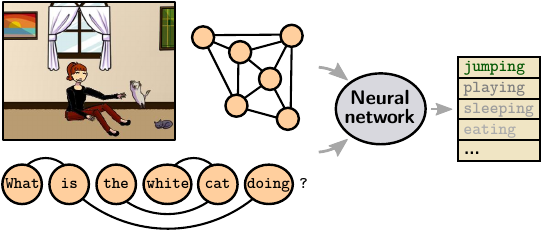}
  \end{center}
  \label{fig:teaser}
  \vspace{-6pt}
  \caption{We encode the input scene as a graph representing the objects and their spatial arrangement, and the input question as a graph representing words and their syntactic dependencies. A neural network is trained to reason over these representations, and to produce a suitable answer as a prediction over an output vocabulary.}
  \vspace{-8pt}
\end{figure}

Multiple datasets for VQA have been introduced with either real \cite{antol2015vqa,krishnavisualgenome,malinowski2014multi,ren2015image,zhu2015visual7w} or synthetic images \cite{antol2015vqa,zhang2015balanced}. Our experiments uses the latter, being based on clip art or ``cartoon'' images created by humans to depict realistic scenes (they are usually referred to as ``abstract scenes'', despite this being a misnomer). Our experiments focus on this dataset of clip art scenes as they allow to focus on semantic reasoning and vision-language interactions, in isolation from the performance of visual recognition (see examples in \fig\ref{fig:qualitative}). 
They also allow the manipulation of the image data so as to better illuminate algorithm performance. A particularly attractive VQA dataset was introduced in \cite{zhang2015balanced} by selecting only the questions with binary answers (\eg yes/no) and pairing each (synthetic) image with a minimally-different complementary version that elicits the opposite (no/yes) answer (see examples in \fig\ref{fig:qualitative}, bottom rows). This strongly contrasts with other VQA datasets of real images, where a correct answer is often obvious without looking at the image, by relying on systematic regularities of frequent questions and answers \cite{antol2015vqa,zhang2015balanced}. 
Performance improvements reported on such datasets are difficult to interpret as actual progress in scene understanding and reasoning as they might similarly be taken to represent a better modeling of the language prior of the dataset.
This hampers, or at best obscures, progress toward the greater goal of general VQA. In our view, and despite obvious limitations of synthetic images, improvements on the aforementioned ``balanced'' dataset constitute an illuminating measure of progress in scene-understanding, because a language model alone cannot perform better than chance on this data.

\paragraph{Challenges}
The questions in the clip-art dataset vary greatly in their complexity. Some can be directly answered from observations of visual elements, \eg \textit{Is there a dog in the room ?}, or \textit{Is the weather good ?}. Others require relating multiple facts or understanding complex actions, \eg \textit{Is the boy going to catch the ball?}, or \textit{Is it winter?}. An additional challenge, which affects all VQA datasets, is the sparsity of the training data. Even a large number of training questions (almost 25,000 for the clip art scenes of \cite{antol2015vqa}) cannot possibly cover the combinatorial diversity of possible objects and concepts. Adding to this challenge, most methods for VQA process the question through a recurrent neural network (such as an LSTM) trained from scratch solely on the training questions.

\paragraph{Language representation}
The above reasons motivate us to take advantage of the extensive existing work in the natural language community to aid processing the questions. First, we identify the syntactic structure of the question using a dependency parser \cite{demarneffe2008parser}. This produces a graph representation of the question in which each node represents a word and each edge a particular type of dependency (\eg \textit{determiner}, \textit{nominal subject}, \textit{direct object}, \etc). Second, we associate each word (node) with a vector embedding pretrained on large corpora of text data \cite{pennington2014glove}. This embedding maps the words to a space in which distances are semantically meaningful. Consequently, this essentially regularizes the remainder of the network to share learned concepts among related words and synonyms. This particularly helps in dealing with rare words, and also allows questions to include words absent from the training questions/answers. Note that this pretraining and \textit{ad hoc} processing of the language part mimics a practice common for the image part, in which visual features are usually obtained from a fixed CNN, itself pretrained on a larger dataset and with a different (supervised classification) objective.

\paragraph{Scene representation}
Each object in the scene corresponds to a node in the scene graph, which has an associated feature vector describing its appearance. The graph is fully connected, with each edge representing the relative position of the objects in the image. 

\paragraph{Applying Neural Networks to graphs} The two graph representations feed into a deep neural network that we will describe in \sect\ref{sec:architecture}. The advantage of this approach with text- and scene-graphs, rather than more typical representations, is that the graphs can capture relationships between words and between objects which are of semantic significance.  This enables the GNN to exploit (1)~the \textit{unordered} nature of scene elements (the objects in particular) and (2)~the \emph{semantic relationships} between elements (and the grammatical relationships between words in particular). This contrasts with the typical approach of representing the image with CNN activations (which are sensitive to individual object locations but less so to relative position) and the processing words of the question serially with an RNN (despite the fact that grammatical structure is very non-linear). The graph representation ignores the order in which elements are processed, but instead represents the relationships between different elements using different edge types. Our network uses multiple layers that iterate over the features associated with every node, then ultimately identifies a soft matching between nodes from the two graphs. This matching reflects the correspondences between the words in the question and the objects in the image. The features of the matched nodes then feed into a classifier to infer the answer to the question (\fig\ref{fig:teaser}).

\vspace{1ex}
\noindent
The main contributions of this paper are four-fold.
\begin{enumerate}[topsep=2pt,itemsep=-1ex,partopsep=1ex,parsep=1ex,label={\arabic*)},leftmargin=3.0ex]
\item We describe how to use graph representations of scene and question for VQA, and a neural network capable of processing these representations to infer an answer.
\item We show how to make use of an off-the-shelf language parsing tool by generating a graph representation of text that captures grammatical relationships, and by making this information accessible to the VQA model. This representation uses a pre-trained word embedding to form node features, and encodes syntactic dependencies between words as edge features.
\item We train the proposed model on the VQA ``abstract scenes'' benchmark \cite{antol2015vqa} and demonstrate its efficacy by raising the state-of-the-art accuracy from 71.2\% to 74.4\% in the multiple-choice setting. On the ``balanced'' version of the dataset, we raise the accuracy from 34.7\% to 39.1\% in the hardest setting (requiring a correct answer over \textit{pairs} of scenes).
\item We evaluate the uncertainty in the model by presenting --~for the first time on the task of VQA~-- precision/recall curves of predicted answers. Those curves provide more insight than the single accuracy metric and show that the uncertainty estimated by the model about its predictions correlates with the ambiguity of the human-provided ground truth.
\end{enumerate}

\section{Related work}

\begin{figure*}[t]
  \vspace{-3pt}
  \includegraphics[width=.95\linewidth]{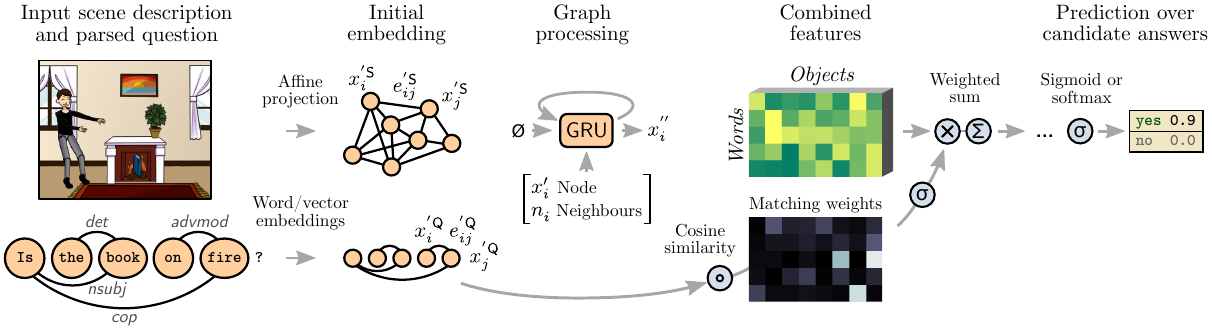}
  \vspace{-1pt}
  \caption{Architecture of the proposed neural network. The input is provided as a description of the scene (a list of objects with their visual characteristics) and a parsed question (words with their syntactic relations). The scene-graph contains a node with a feature vector for each object, and edge features that represent their spatial relationships. The question-graph reflects the parse tree of the question, with a word embedding for each node, and a vector embedding of types of syntactic dependencies for edges. A recurrent unit (GRU) is associated with each node of both graphs. Over multiple iterations, the GRU updates a representation of each node that integrates context from its neighbours within the graph. Features of all objects and all words are combined (concatenated) pairwise, and they are weighted with a form of attention. That effectively matches elements between the question and the scene. The weighted sum of features is passed through a final classifier that predicts scores over a fixed set of candidate answers.}
  \label{fig:network}
  \vspace{-10pt}
\end{figure*}

The task of {visual question answering} has received increasing interest since the seminal paper of Antol \etal \cite{antol2015vqa}. Most recent methods are based on the idea of a \textbf{joint embedding} of the image and the question using a deep neural network. The image is passed through a convolutional neural network (CNN) pretrained for image classification, from which intermediate features are extracted to describe the image. The question is typically passed through a recurrent neural network (RNN) such as an LSTM, which produces a fixed-size vector representing the sequence of words. These two representations are mapped to a joint space by one or several non-linear layers. They can then be fed into a classifier over an output vocabulary, predicting the final answer. Most recent papers on VQA propose improvements and variations on this basic idea. Consult \cite{wu2016survey} for a survey.

A major improvement to the basic method is to use an \textbf{attention mechanism} \cite{zhu2015visual7w,xu2015ask,Chen2015ABC,Jiang2015compositional,andreas2015deep,yang2015stacked}. It models interactions between specific parts of the inputs (image and question) depending on their actual contents. The visual input is then typically represented a spatial feature map, instead of holistic, image-wide features. The feature map is used with the question to determine spatial weights that reflect the most relevant regions of the image. Our approach uses a similar weighting operation, which, with our graph representation, we equate to a subgraph matching. Graph nodes representing question words are associated with graph nodes representing scene objects and vice versa. Similarly, the co-attention model of Lu \etal \cite{lu2016hierarchical} determines attention weights on both image regions and question words. Their best-performing approach proceeds in a sequential manner, starting with question-guided visual attention followed by image-guided question attention. In our case, we found that a joint, one-pass version performs better.

A major contribution of our model is to use \textbf{structured representations} of the input scene and the question. This contrasts with typical CNN and RNN models which are limited to spatial feature maps and sequences of words respectively. The dynamic memory networks (DMN), applied to VQA in \cite{xiong2016} also maintain a set-like representation of the input. As in our model, the DMN models interactions between different parts of the input. Our method can additionally take, as input, features characterizing  arbitrary relations between parts of the input (the edge features in our graphs). This specifically allows making use of syntactic dependencies between words after pre-parsing the question.

Most VQA systems are trained end-to-end from questions and images to answers, with the exception of the visual feature extractor, which is typically a CNN pretrained for image classification. For the \textbf{language processing} part, some methods address the the semantic aspect with word embeddings pretrained on a language modeling task (\eg~\cite{shih2015look,fukui2016multimodal}). The syntactic relationships between the words in the question are typically overlooked, however. In \cite{zhang2015balanced}, hand-designed rules serve to identify primary and secondary objects of the questions. In the Neural Module Networks \cite{andreas2015deep,andreas2016learning}, the question is processed by a dependency parser, and fragments of the parse, selected with \textit{ad hoc} fixed rules are associated with modules, are assembled into a full neural network. In contrast, our method is trained to make direct use of the output of a syntactic parser.

\textbf{Neural networks on graphs} have received significant attention recently \cite{duvenaud2015molecular,jain2016graphs,li2015graph}. The approach most similar to ours is the Gated Graph Sequence Neural Network \cite{li2015graph}, which associate a gated recurrent unit (GRU \cite{cho2014learning}) to each node, and updates the feature vector of each node by iteratively passing messages between neighbours. Also related is the work of Vinyals \etal~\cite{vinyals2015sets} for embedding a set into fixed-size vector, invariant to the order of its elements. They do so by feeding the entire set through a recurrent unit multiple times. Each iteration uses an attention mechanism to focus on different parts of the set. Our formulation similarly incorporates information from neighbours into each node feature over multiple iterations, but we did not find any advantage in using an attention mechanism within the recurrent unit.


\section{Graph representation of scenes and questions}
\vspace{-4pt}

The input data for each training or test instance is a question, and a parameterized description of contents of the scene. The question is processed with the Stanford dependency parser \cite{demarneffe2008parser}, which outputs the following.
\begin{enumerate}[topsep=0pt,itemsep=-1ex,partopsep=1ex,parsep=1.0ex,label={\tiny$\bullet$},leftmargin=2.0ex]
\vspace{1pt}
\item A set of $N^\symbQ$ words that constitute the nodes of the question graph. Each word is represented by its index in the input vocabulary, a token $x_i^\symbQ \in \mathbb{Z}$ ($i\in1..N^\symbQ$).
\vspace{1pt}
\item A set of pairwise relations between words, which constitute the edges of our graph. An edge between words $i$ and $j$ is represented by $e_{ij}^\symbQ \in \mathbb{Z}$, an index among the possible types of dependencies.
\end{enumerate}
\vspace{1pt}
The dataset provides the following information about the image
\begin{enumerate}[topsep=0pt,itemsep=-1ex,partopsep=1ex,parsep=1.0ex,label={\tiny$\bullet$},leftmargin=2.0ex]
\vspace{1pt}
\item A set of $N^\symbS$ objects that constitute the nodes of the scene graph. Each node is represented by a vector $x_i^{\symbS} \in \mathbb{R}^C$ of visual features ($i\in1..N^\symbS$). Please refer to the supplementary material for implementation details.
\vspace{1pt}
\item A set of pairwise relations between all objects. They form the edges of a fully-connected graph of the scene. The edge between objects $i$ and $j$ is represented by a vector $e_{ij}^\symbS \in \mathbb{R}^D$ that encodes relative spatial relationships (see supp. mat.).
\end{enumerate}
\vspace{1pt}
Our experiments are carried out on datasets of clip art scenes, in which descriptions of the scenes are provided in the form of lists of objects with their visual features. The method is equally applicable to real images, with the object list replaced by candidate object detections. Our experiments on clip art allows the effect of the proposed method to be isolated from the performance of the object detector.
Please refer to the supplementary material for implementation details.

The features of all nodes and edges are projected to a vector space $\mathbb{R}^H$ of common dimension (typically $H$=300). 
The question nodes and edges use vector embeddings implemented as look-up tables, and the scene nodes and edges use affine projections:
\abovedisplayskip=3pt
\belowdisplayskip=3pt
\begin{flalign}
  & x_i^{'\symbQ} = W_1\big[ x^\symbQ_i 
	\big] ~~~~~~\;\; e_{ij}^{'\symbQ} = W_2\big[ e_{ij}^\symbQ \big] \\
  & x_i^{'\symbS} = W_3 x_i^\symbS + b_3 ~~~~~ e_{ij}^{'\symbS} = W_4 e_{ij}^\symbS + b_4
\end{flalign}
with $W_1$ the word embedding (usually pretrained, see supplementary material), $W_2$ the embedding of dependencies, $W_3 \in \mathbb{R}^{h \times c}$ and $W_4 \in \mathbb{R}^{h \times d}$ weight matrices, and $b_3 \in \mathbb{R}^c$ and $b_4 \in \mathbb{R}^d$ biases.

\section{Processing graphs with neural networks}
\label{sec:architecture}

We now describe a deep neural network suitable for processing the question and scene graphs to infer an answer. See \fig\ref{fig:network} for an overview.

The two graphs representing the question and the scene are processed independently in a recurrent architecture. We drop the exponents $\symbS$ and $\symbQ$ for this paragraph as the same procedure applies to both graphs.
Each node $x_i$ is associated with a gated recurrent unit (GRU \cite{cho2014learning}) and processed over a fixed number $T$ of iterations (typically $T$=4):
\begin{flalign}
  \label{eq:gru}
  & h^0_i = 0\\ 
	& n_i = \pool\nolimits_j (\,e'_{ij} \circ x'_j\,)\\
	& h^t_i = \gru \hspace{-.2ex} \big( \, h^{t-1}_i, \; [ x'_i \,;\, n_i ] \, \big) ~~~~~~t \in [1,T].
\end{flalign}
Square brackets with a semicolon represent a concatenation of vectors, and $\circ$ the Hadamard (element-wise) product. The final state of the GRU is used as the new representation of the nodes: $x''_i~=~h^T_i$. The $\pool$ operation transforms features from a variable number of neighbours (\ie connected nodes) to a fixed-size representation. Any commutative operation can be used (\eg sum, maximum). In our implementation, we found  the best performance with the average function, taking care of averaging over the variable number of connected neighbours. An intuitive interpretation of the recurrent processing is to progressively integrate context information from connected neighbours into each node's own representation. 
A node corresponding to the word 'ball', for instance, might thus incorporate the fact that the associated adjective is 'red'. Our formulation is similar but slightly different from the gated graph networks \cite{li2015graph}, as the propagation of information in our model is limited to the first order. 
Note that our graphs are typically densely connected.

We now introduce a form of attention into the model, which constitutes an essential part of the model. The motivation is two-fold: (1)~to identify parts of the input data most relevant to produce the answer and (2)~to align specific words in the question with particular elements of the scene. Practically, we estimate the relevance of each possible pairwise combination of words and objects. More precisely, we compute scalar ``matching weights'' between node sets $\{x_i^{'\symbQ}\}$ and $\{x_i^{'\symbS}\}$. These weights are comparable to the ``attention weights'' in other models (\eg~\cite{lu2016hierarchical}). Therefore, $\forall ~~i\in1..N^\symbQ, j\in1..N^\symbS$:
\begin{gather} \label{eq:weights}
  a_{ij} = \sigma \Bigg( W_5 \Big( \frac{x_i^{'\symbQ}}{\norm{x_i^{'\symbQ}}} \circ \frac{x_j^{'\symbS}}{\norm{x_j^{'\symbS}}} \Big) + b_5 \Bigg)
\end{gather}
where $W_5 \in \mathbb{R}^{1 \times h}$  and $b_5 \in \mathbb{R}$ 
are learned weights and biases, and $\sigma$ the logistic function that introduces a non-linearity and bounds the weights to $(0,1)$. The formulation is similar to a cosine similarity with learned weights on the feature dimensions. Note that the weights are computed using the initial embedding of the node features (pre-GRU). We apply the scalar weights $a_{ij}$ to the corresponding pairwise combinations of question and scene features, thereby focusing and giving more importance to the matched pairs~(\eq\ref{eq:out1}). We sum the weighted features over the scene elements~(\eq\ref{eq:out2}) then over the question elements~(\eq\ref{eq:out3}), interleaving the sums with affine projections and non-linearities to obtain a final prediction:
\begin{flalign}
  & y_{ij} ~=~ a_{ij} \, . \, [ x^{''\symbQ}_i \,;\, x^{''\symbS}_j ] \label{eq:out1} \\
  & y'_i \, ~=~ f \big( W_6 \textstyle\sum_j^{N^\symbS} y_{ij} + b_6 \big) \label{eq:out2} \\
  & y'' ~=~ f' \big( W_7 \textstyle \sum_i^{N^\symbQ} y'_i + b_7 \big) \label{eq:out3}
\end{flalign}
with $W_6$, $W_7$, $b_6$, $b_7$ learned weights and biases, $f$ a ReLU, and $f'$ a softmax or a logistic function (see experiments, \sect\ref{sec:pr}). The summations over the scene elements and question elements is a form of pooling that brings the variable number of features (due to the variable number of words and objects in the input) to a fixed-size output. The final output vector $y'' \in \mathbb{R}^T$ contains scores for the possible answers, and has a number of dimensions equal to 2 for the binary questions of the ``balanced'' dataset, or to the number of all candidate answers in the ``abstract scenes'' dataset. The candidate answers are those appearing at least $5$ times in the training set (see supplementary material for details).

\section{Evaluation}
\label{sec:evaluation}

\paragraph{Datasets}
Our evaluation uses two datasets: the original ``abstract scenes'' from Antol \etal \cite{antol2015vqa} and its ``balanced'' extension from \cite{zhang2015balanced}. They both contain scenes created by humans in a drag-and-drop interface for arranging clip art objects and figures. The original dataset contains $20k/10k/20k$ scenes (for training$/$validation$/$test respectively) and $60k/30k/60k$ questions, each with 10 human-provided ground-truth answers. Questions are categorized based on the type of the correct answer into \textit{yes/no}, \textit{number}, and \textit{other}, but the same method is used for all categories, the type of the test questions being unknown. The ``balanced'' version of the dataset contains only the subset of questions which have binary (yes/no) answers and, in addition, complementary scenes created to elicit the opposite answer to each question. This is significant because guessing the modal answer from the training set will the succeed only half of the time (slightly more than $50\%$ in practice because of disagreement between annotators) and give $0\%$ accuracy over complementary pairs. This contrasts with other VQA datasets where blind guessing can be very effective. The pairs of complementary scenes also typically differ by only one or two objects being displaced, removed, or slightly modified (see examples in \fig\ref{fig:qualitative}, bottom rows). This makes the questions very challenging by requiring to take into account subtle details of the scenes.

\paragraph{Metrics}
The main metric is the average ``VQA score'' \cite{antol2015vqa}, which is a soft accuracy that takes into account variability of ground truth answers from multiple human annotators. Let us refer to a test question by an index $q=1..M$, and to each possible answer in the output vocabulary by an index $a$. The \textbf{ground truth score} $s(q,a)=1.0$ if the answer $a$ was provided by $m\nospace\geq\nospace3$ annotators. Otherwise, $s(q,a)=m/3$\footnote{Ground truth scores are also averaged in a 10--\textit{choose}--9 manner~\cite{antol2015vqa}.}. Our method outputs a \textbf{predicted score} $\hat{s}(q,a)$ for each question and answer ($y''$ in \eq\ref{eq:out3}) and the overall accuracy is the average ground truth score of the highest prediction per question, \ie $\frac{1}{M} \sum_q^{M} s(q,\argmax_a \hat{s}(q,a) )$.

It has been argued that the ``balanced'' dataset can better evaluate a method's level of visual understanding than other datasets, because it is less susceptible to the use of language priors and dataset regularities (\ie guessing from the question\cite{zhang2015balanced}). Our initial experiments confirmed that the performances of various algorithms on the balanced dataset were indeed better separated, and we used it for our ablative analysis. We also focus on the hardest evaluation setting \cite{zhang2015balanced}, which measures the accuracy over \emph{pairs} of complementary scenes. This is the only metric in which blind models (guessing from the question) obtain null accuracy. This setting also does not consider pairs of test scenes deemed ambiguous because of disagreement between annotators. Each test scene is still evaluated independently however, so the model is unable to increase performance by forcing opposite answers to pairs of questions. The metric is then a standard ``hard'' accuracy, \ie all ground truth scores $s(i,j)\in\{0,1\}$. Please refer to the supplementary material for additional details.

\begin{figure*}[t]
  \begin{center}
  \includegraphics[width=0.29\linewidth]{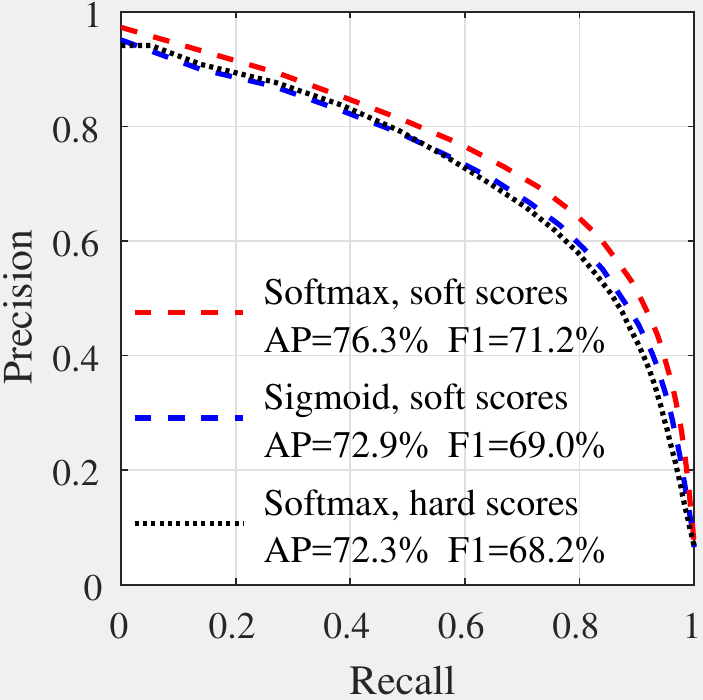}\hspace{2.0em}
  \includegraphics[width=0.29\linewidth]{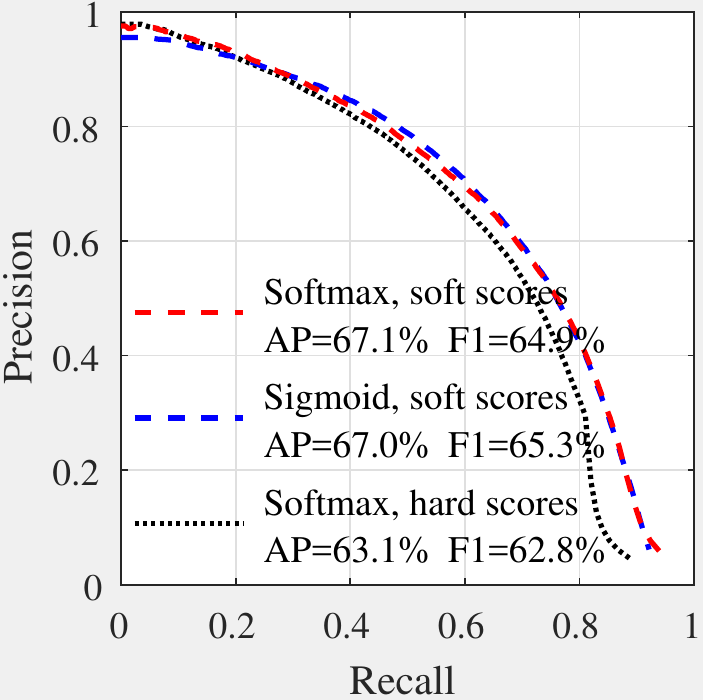}\hspace{2.0em}
  \includegraphics[width=0.29\linewidth]{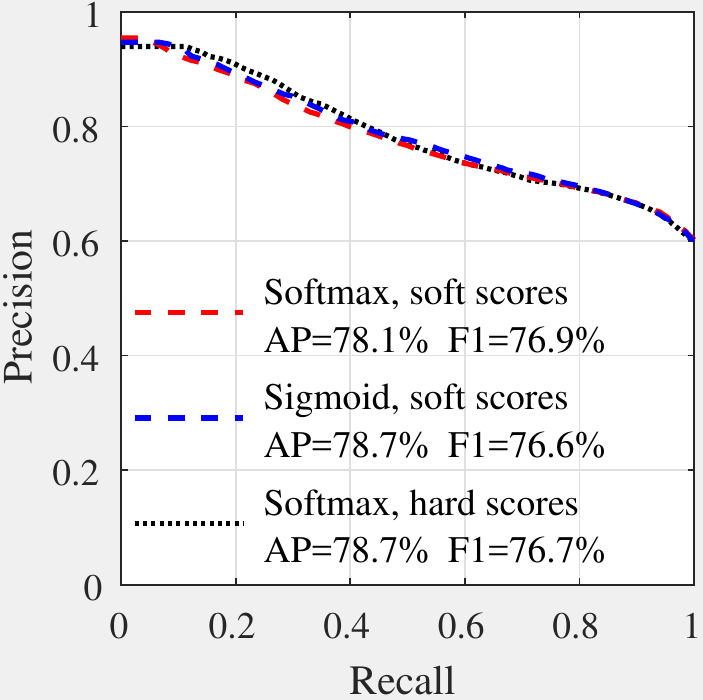}
  \end{center}
  \caption{Precision/recall on the ``abstract scenes'' (\textbf{left}: multiple choice, \textbf{middle}: open-ended) and ``balanced'' datasets (\textbf{right}). The scores assigned by the model to predicted answers is a reliable measure of its certainty: a strict threshold (low recall) filters out incorrect answers and produces a very high precision. On the ``abstract scenes'' dataset (left and middle), a slight advantage is brought by training for soft target scores that capture ambiguities in the human-provided ground truth.}
  \label{fig:pr}
\end{figure*}

\subsection{Evaluation on the ``balanced'' dataset}

We compare our method against the three models proposed in \cite{zhang2015balanced}. They all use an ensemble of models exploiting either an LSTM for processing the question, or an elaborate set of hand-designed rules to identify two objects as the focus of the question. The visual features in the three models are respectively empty (blind model), global (scene-wide), or focused on the two objects identified from the question. These models are specifically designed for binary questions, whereas ours is generally applicable. Nevertheless, we obtain significantly better accuracy than all three (\tab\ref{tab:balanced}). Differences in performance are mostly visible in the ``pairs'' setting, which we believe is more reliable as it discards ambiguous test questions on which human annotators disagreed.

During training, we take care to keep pairs of complementary scenes together when forming mini-batches. This has a significant positive effect on the stability of the optimization. Interestingly, we did not notice any tendency toward overfitting when training on balanced scenes. We hypothesize that the pairs of complementary scenes have a strong regularizing effect that force the learned model to focus on relevant details of the scenes. In \fig\ref{fig:qualitative} (and in the supplementary material), we visualize the matching weights between question words and scene objects (\eq\ref{eq:weights}). As expected, these tend to be larger between semantically related elements (\eg daytime$\leftrightarrow$sun, dog$\leftrightarrow$puppy, boy$\leftrightarrow$human) although some are more difficult to interpret.

Our best performance of about $39\%$ is still low in absolute terms, which is understandable from the wide range of concepts involved in the questions (see examples in \fig\ref{fig:qualitative} and in the supplementary material). It seems unlikely that these concepts could be learned from training question/answers alone, and we suggest that any further significant improvement in performance will require external sources of information at training and/or test time.

\paragraph{Ablative evaluation}
We evaluated variants of our model to measure the impact of various design choices (see numbered rows in \tab\ref{tab:balanced}). On the \textbf{question side}, we evaluate (row 1) our graph approach without syntactic parsing, building question graphs with only two types of edges, \textit{previous}/\textit{next} and linking consecutive nodes. This shows the advantage of using the graph method together with syntactic parsing. Optimizing the word embeddings from scratch (row 2) rather than from pretrained Glove vectors \cite{pennington2014glove} produces a significant drop in performance. On the \textbf{scene side}, we removed the edge features (row 3) by setting $e_{ij}^\symbS=1$. It confirms that the model makes use of the spatial relations between objects encoded by the edges of the graph. In rows 4--6, we disabled the recurrent graph processing ($x''_i=x'_i$) for the either the question, the scene, or both. We finally tested the model with uniform matching weights ($a_{ij}=1$, row 10). As expected, it performed poorly. Our weights act similarly to the attention mechanisms in other models (\eg\cite{zhu2015visual7w,xu2015ask,Chen2015ABC,Jiang2015compositional,yang2015stacked}) and our observations confirm that such mechanisms are crucial for good performance.


\renewcommand{\arraystretch}{1.2}

\begin{table}
  \vspace{-4pt}
  \small
  \renewcommand{\tabcolsep}{0.1mm}
  \renewcommand{\arraystretch}{1.3}
  \begin{center}
  \begin{tabular*}{\linewidth}{@{}>{}l@{\extracolsep{\fill}}*{2}{c}@{}}
  ~ & Avg. score & Avg. accuracy \\
  Method & over scenes & over pairs \\
  \hline
  Zhang \etal \cite{zhang2015balanced} blind & 63.33 & 0.00 \\
  ~~with global image features & 71.03 & 23.13 \\
  ~~with attention-based image features & 74.65 & 34.73 \\
  \hline
  \textbf{Graph VQA} (full model) & \textbf{74.94} & \textbf{39.1} \\
  \end{tabular*}

  \begin{tabular*}{\linewidth}{@{}>{}l@{\extracolsep{\fill}}*{2}{c}@{}}
  \scriptsize{(1)}\small~~Question: no parsing (graph with previous/next edges) & 37.9\\
  \scriptsize{(2)}\small~~Question: word embedding not pretrained & 33.8\\
  \scriptsize{(3)}\small~~Scene: no edge features ($e_{ij}^{'\symbS}$=$1$) & 36.8\\
  \scriptsize{(4)}\small~~Graph processing: disabled for question ($x^{''\symbQ}_i$=$x^{'\symbS}_i$) & 37.1\\
  \scriptsize{(5)}\small~~Graph processing: disabled for scene ($x^{''\symbS}_i$=$x^{'\symbQ}_i$) & 37.0\\
  \scriptsize{(6)}\small~~Graph processing: disabled for question/scene & 35.7\\
  \scriptsize{(7)}\small~~Graph processing: only 1 iteration for question ($T^\symbQ$=1) & 39.0\\
  \scriptsize{(8)}\small~~Graph processing: only 1 iteration for scene ($T^\symbS$=1) & 37.9\\
  \scriptsize{(9)}\small~~Graph processing: only 1 iteration for question/scene & 39.1\\
  \scriptsize{(10)}\small~~Uniform matching weights ($a_{ij}$=$1$) & 24.4\\
  \hline
  \end{tabular*}
  \end{center}
  \caption{Results on the test set of the ``balanced'' dataset \cite{zhang2015balanced} (in percents , using balanced versions of both training and test sets). Numbered rows report accuracy over pairs of complementary scenes for ablated versions of our method.}
  \label{tab:balanced}
\end{table}

\begin{figure}[t]
  \begin{center}
  \includegraphics[width=0.95\linewidth]{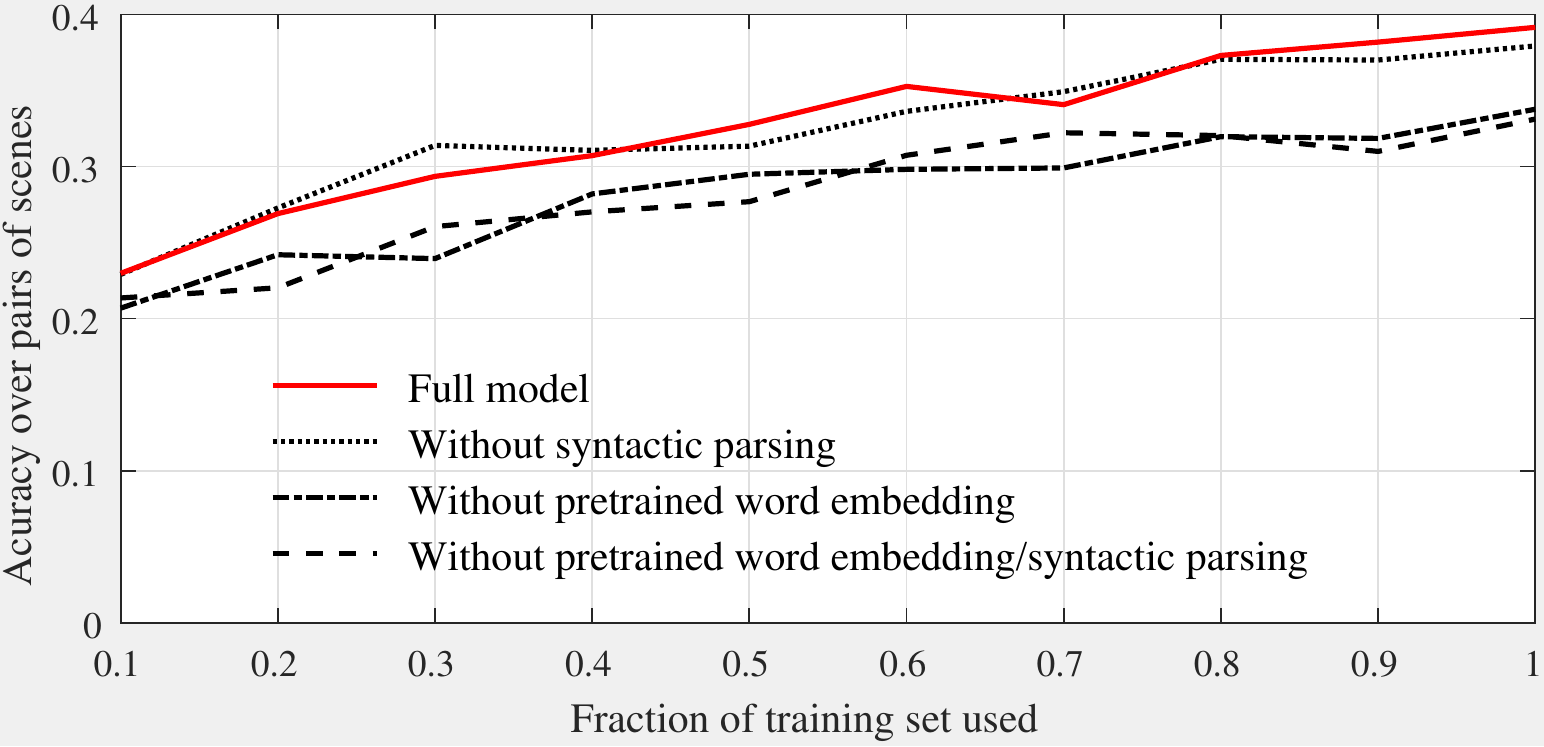}
  \end{center}
  \vspace{-7pt}
  \caption{Impact of the amount of training data on performance (accuracy over pairs on the ``balanced'' dataset). Language preprocessing always improve generalization: pre-parsing and pretrained word embeddings both have a positive impact individually, and their effects are complementary to each other.}
  \label{fig:limitedTrainingData}
  \vspace{-3pt}
\end{figure}

\paragraph{Precision/recall}
\label{sec:pr}
We are interested in assessing the confidence of our model in its predicted answers. Most existing VQA methods treat the answering as a hard classification over candidate answers, and almost all reported results consist of a single accuracy metric. To provide more insight, we produce precision/recall curves for predicted answers. A precision/recall point $(p,r)$ is obtained by setting a threshold $t$ on predicted scores such that
\begin{flalign}
  & p = \frac{        \sum_{i,j} \mathbbm{1}\big( \hat{s}(i,j) \shg t \big)  \;  s(i,j)       }{\sum_{i,j} \mathbbm{1}( \hat{s}(i,j) \shg t) }  \\
  & r = \frac{        \sum_{i,j} \mathbbm{1}\big( \hat{s}(i,j) \shg t \big)  \;  s(i,j)       }{\sum_{i,j} s(i,j)}
\end{flalign}
where $\mathbbm{1}(\cdot)$ is the $0/1$ indicator function. We plot precision/recall curves in \fig\ref{fig:pr} for both datasets\footnote{The ``abstract scenes'' test set is not available publicly, and precision/recall can only be provided on its validation set.}. The predicted score proves to be a reliable indicator of the model confidence, as a low threshold can achieve near-perfect accuracy (\fig\ref{fig:pr}, left and middle) by filtering out harder and/or ambiguous test cases.

We compare models trained with either a softmax or a sigmoid as the final non-linearity (\eq\ref{eq:out3}). The common practice is to train the softmax for a hard classification objective, using a cross-entropy loss and the answer of highest ground truth score as the target. In an attempt to make better use of the multiple human-provided answers, we propose to use the \emph{soft} ground truth scores as the target with a logarithmic loss. This shows an advantage on the ``abstract scenes'' dataset (\fig\ref{fig:pr}, left and middle). In that dataset, the soft target scores reflect frequent ambiguities in the questions and the scenes, and when synonyms constitute multiple acceptable answers. In those cases, we can avoid the potential confusion induced by a hard classification for one specific answer. The ``balanced'' dataset, by nature, contains almost no such ambiguities, and there is no significant difference between the different training objectives (\fig\ref{fig:pr}, right).

\paragraph{Effect of training set size}
Our motivation for introducing language parsing and pretrained word embeddings is to better generalize the concepts learned from the limited training examples. Words representing semantically close concepts ideally get assigned close word embeddings. Similarly, paraphrases of similar questions should produce parse graphs with more similarities than a simple concatenation of words would reveal (as in the input to traditional LSTMs).

We trained our model with limited subsets of the training data (see \fig\ref{fig:limitedTrainingData}). Unsurprisingly, the performance grows steadily with the amount of training data, which suggests that larger datasets would improve performance. In our opinion however, it seems unlikely that sufficient data, covering all possible concepts, could be collected in the form of question/answer examples. More data can however be brought in with other sources of information and supervision. Our use of parsing and word embeddings is a small step in that direction. Both techniques clearly improve generalization (\fig\ref{fig:limitedTrainingData}). The effect may be particularly visible in our case because of the relatively small number of training examples (about $20$k questions in the ``balanced'' dataset). It is unclear whether huge VQA datasets could ultimately negate this advantage. Future experiments on larger datasets (\eg \cite{krishnavisualgenome}) may answer this question.


\subsection{Evaluation on the ``abstract scenes'' dataset}
We report our results on the original ``abstract scenes'' dataset in \tab\ref{tab:benchmark}. The evaluation is performed on an automated server that does not allow for an extensive ablative analysis. Anecdotally, performance on the validation set corroborates all findings presented above, in particular the strong benefit of pre-parsing, pretrained word embeddings, and graph processing with a GRU. At the time of our submission, our method occupies the top place on the leader board in both the open-ended and multiple choice settings. The advantage over existing method is most pronounced on the binary and the counting questions. Refer to \fig\ref{fig:qualitative} and to the supplementary for visualizations of the results.



\begin{table*}
  \small
  \renewcommand{\tabcolsep}{0.1mm}
  \begin{center}
  \begin{tabularx}{\textwidth}{l c *8{>{\Centering\arraybackslash}X}}
  ~ & \multicolumn{4}{l}{\st{~~~~~~~~~~~~~~~~~~~}~~~~Multiple choice~~~~\st{~~~~~~~~~~~~~~~~~}} & \multicolumn{4}{r}{\st{~~~~~~~~~~~~~~~~~~~~~~~~~}~~~~Open-ended~~~~\st{~~~~~~~~~~~~~~~~~~~~~~~~}} \\
  Method & Overall & Yes/no & Other & Number~~~ & ~~~Overall & Yes/no & Other & Number \\
  \hline
  LSTM blind \cite{antol2015vqa}							& 61.41 & 76.90 & 49.19 & 49.65							& 57.19 & 76.88 & 38.79 & 49.55 \\
  LSTM with global image features \cite{antol2015vqa}~~~~~~~& 69.21 & 77.46 & 66.65 & 52.90							& 65.02 & 77.45 & 56.41 & 52.54 \\
  Zhang \etal \cite{zhang2015balanced} (yes/no only)		& 35.25 & 79.14 & --- & --- 							& 35.25 & 79.14 & --- & ---  \\
  Multimodal residual learning \cite{kim2016multimodal} 	& 67.99 & 79.08 & 61.99 & 52.57 						& 62.56 & 79.10 & 48.90 & 51.60 \\
  U. Tokyo MIL (ensemble) \cite{saito2016dualnet,vqaleaderboard} 	& 71.18 & 79.59 &	67.93 & 56.19 		& 69.73 & 80.70 & \textbf{62.08} & 58.82 \\
  \hline
  \textbf{Graph VQA} (full model) 							& \textbf{74.37} & \textbf{79.74} & \textbf{68.31} & \textbf{74.97} 	& \textbf{70.42} & \textbf{81.26} & 56.28 & \textbf{76.47} \\
  \hline
  \end{tabularx}
  \end{center}
  \vspace{-3pt}
  \caption{Results on the test set of the ``abstract scenes'' dataset (average scores in percents).}
  \label{tab:benchmark}
  \vspace{25pt}
\end{table*}

\begin{figure*}[t]
  \begin{tabularx}{\textwidth}{*{4}{>{\Centering\arraybackslash}X}}
  \renewcommand{\tabcolsep}{0.1mm}
  \renewcommand{\arraystretch}{1.0}
  \footnotesize
  
  \vspace{-12mm}
  \hangBox{
    \includegraphics[width=42mm]{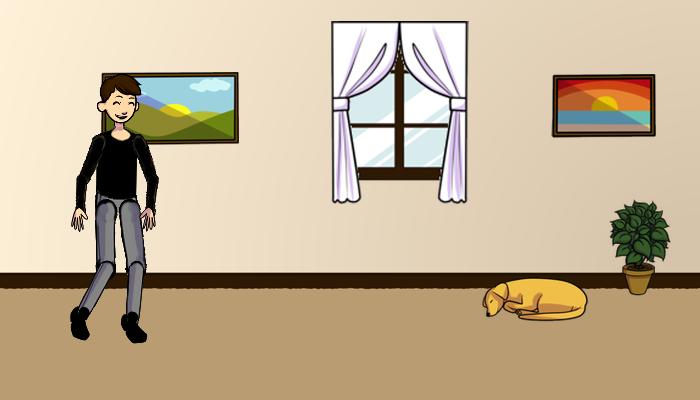}\\
    \includegraphics[width=42mm]{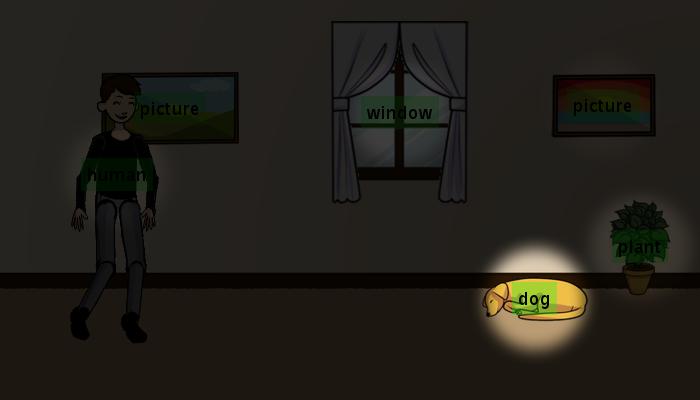}\\
  }&
    \begin{minipage}{\linewidth}\footnotesize\input{"20020.txt"}\end{minipage}
    \vspace{7mm}
    \hangBox{\includegraphics[width=40mm]{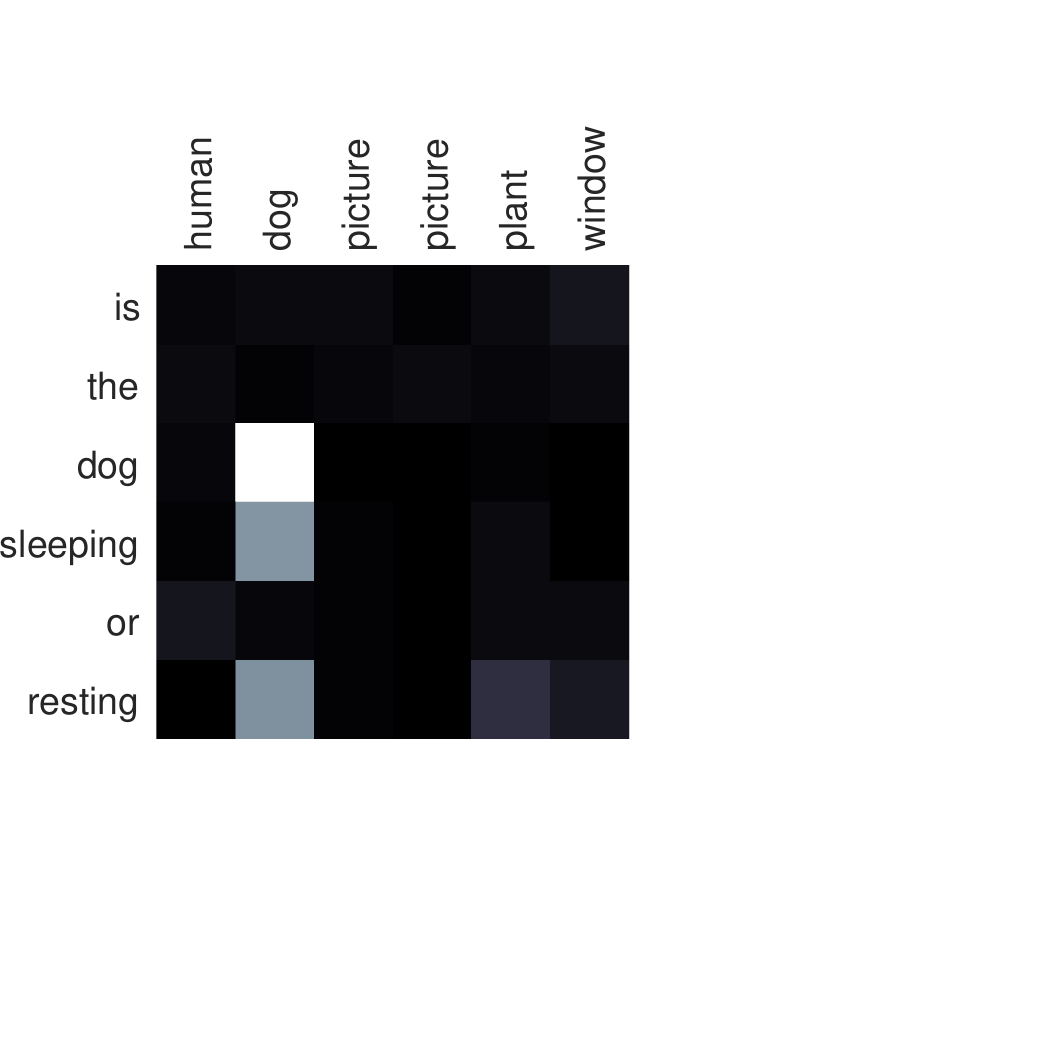}}\vspace{-7mm}
  &
  \vspace{-12mm}
  \hangBox{
    \includegraphics[width=42mm]{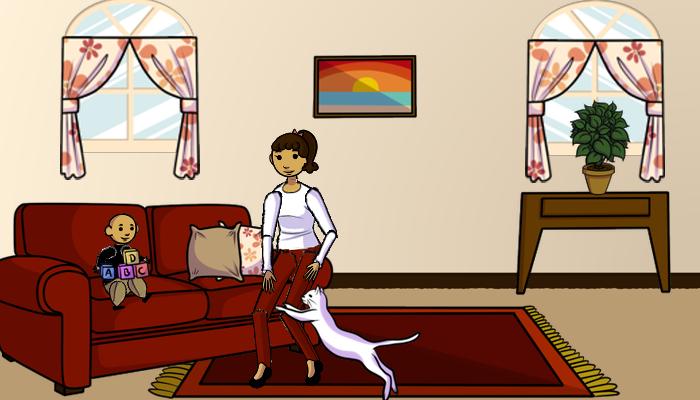}\\
    \includegraphics[width=42mm]{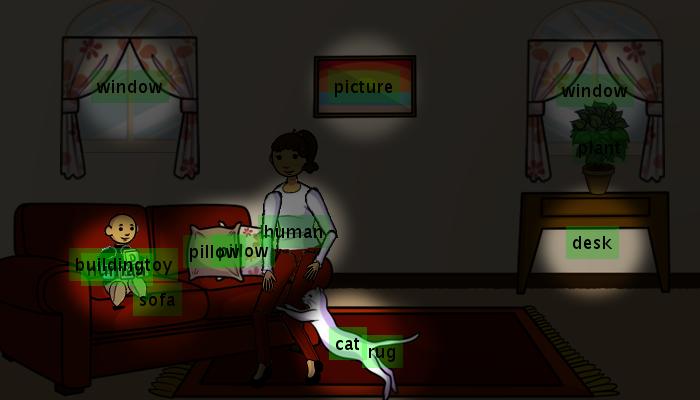}\\
  }&
    \begin{minipage}{\linewidth}\footnotesize\input{"20046.txt"}\end{minipage}
    \vspace{7mm}
    \hangBox{\includegraphics[width=40mm]{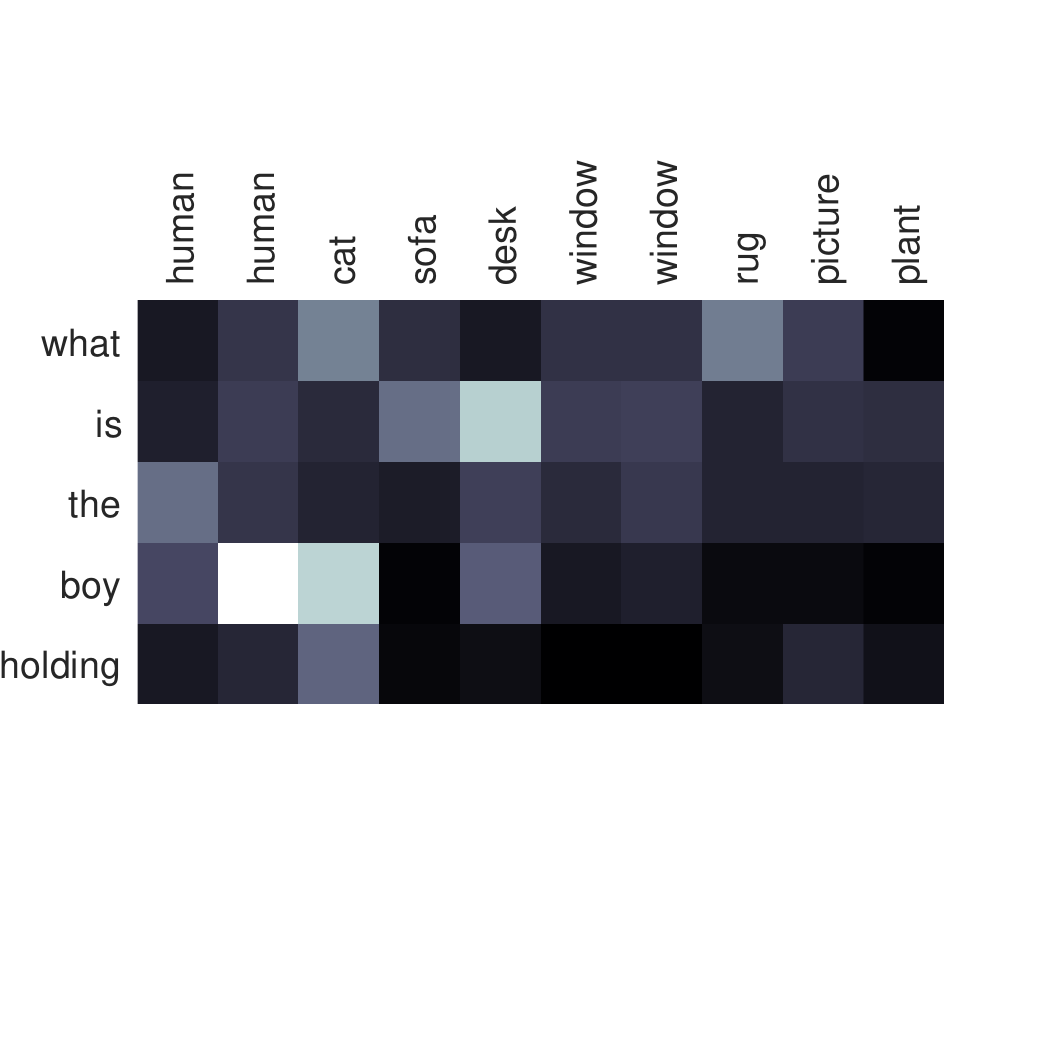}}\vspace{-7mm}
  \\
\\
  
  \vspace{-12mm}
  \hangBox{
    \includegraphics[width=42mm]{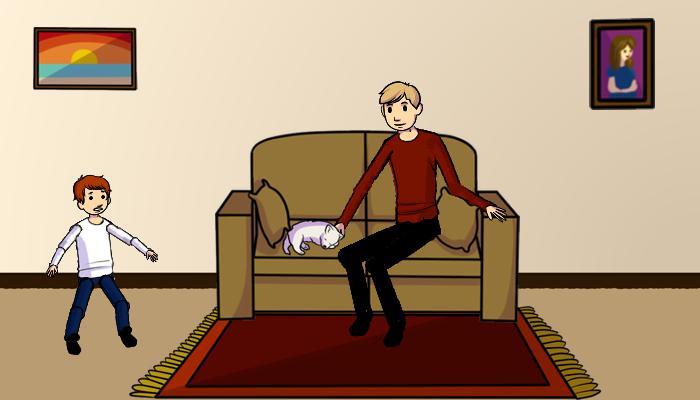}\\
    \includegraphics[width=42mm]{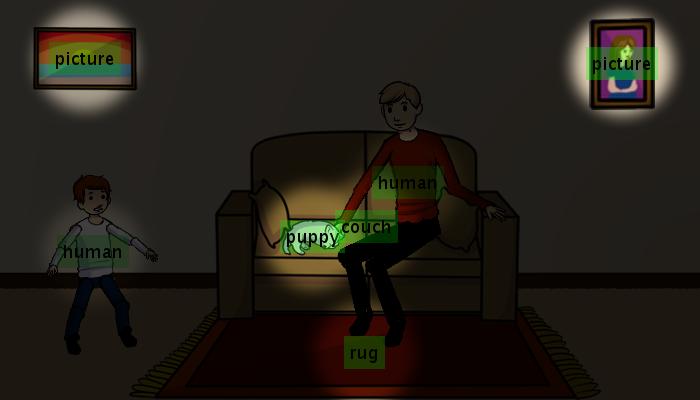}\\
  }&
    \begin{minipage}{\linewidth}\footnotesize\input{"8057-11746.txt"}\end{minipage}
    \vspace{7mm}
    \hangBox{\includegraphics[width=40mm]{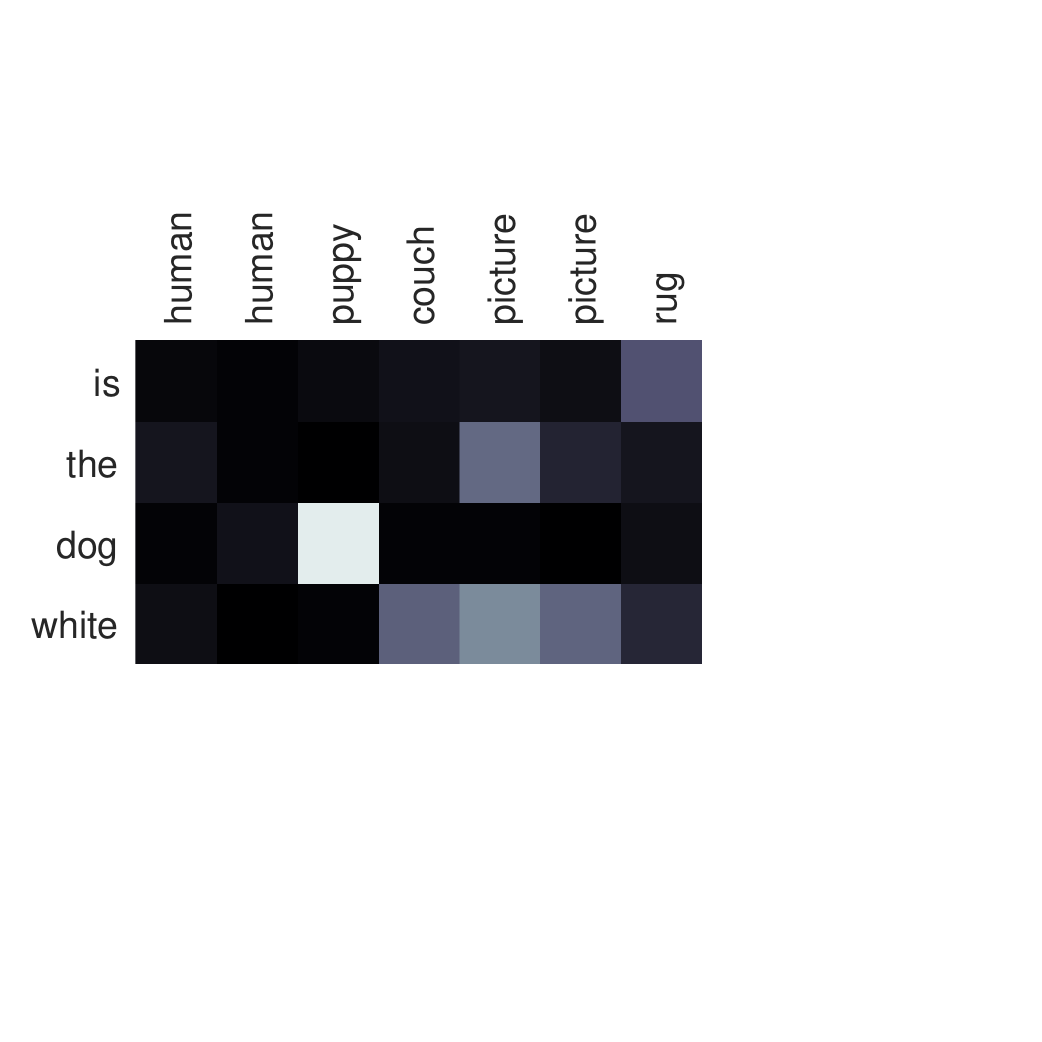}}\vspace{-7mm}
  &
  \vspace{-12mm}
  \hangBox{
    \includegraphics[width=42mm]{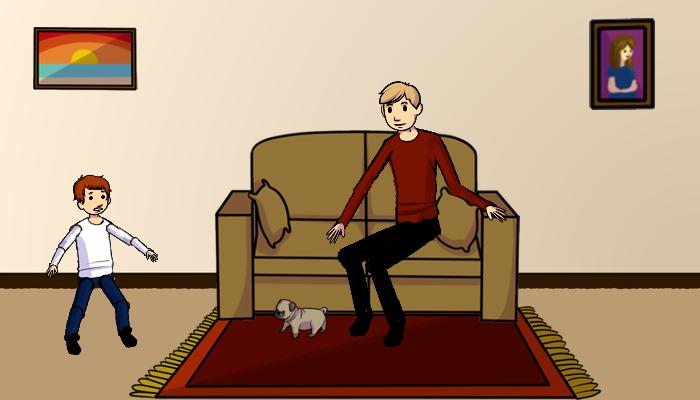}\\
    \includegraphics[width=42mm]{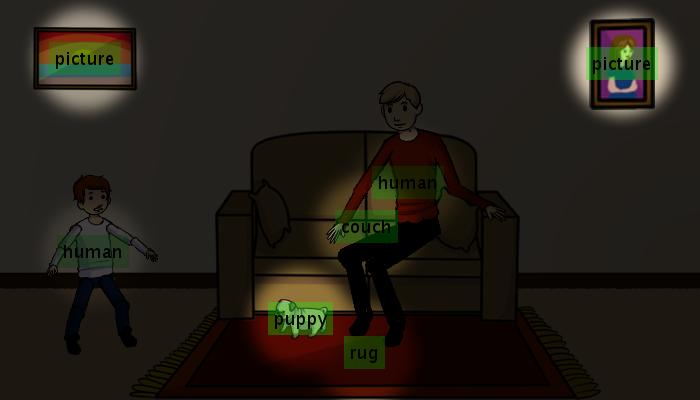}\\
  }&
    \begin{minipage}{\linewidth}\footnotesize\input{"8058-900117462.txt"}\end{minipage}
    \vspace{7mm}
    \hangBox{\includegraphics[width=40mm]{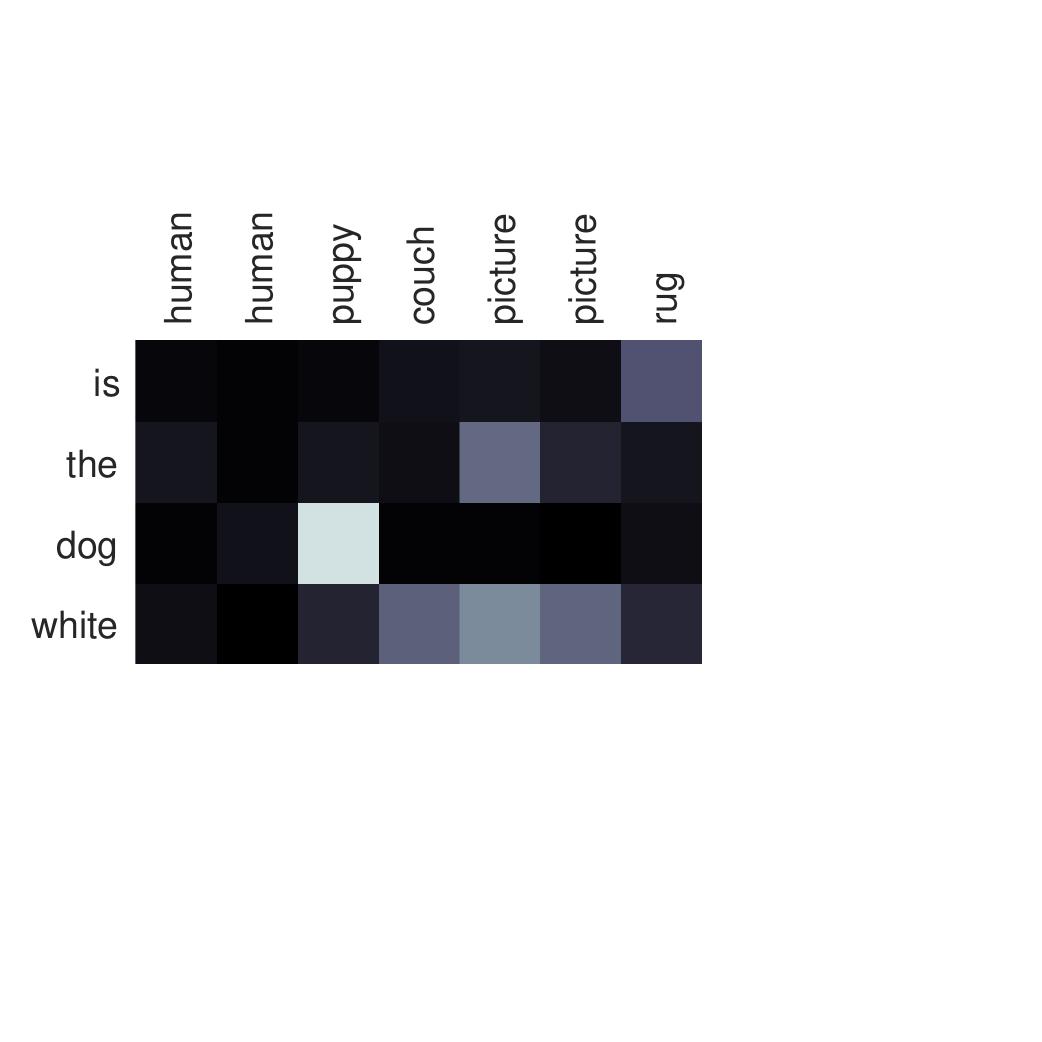}}\vspace{-7mm}
\\
  
  \vspace{-12mm}
  \hangBox{
    \includegraphics[width=42mm]{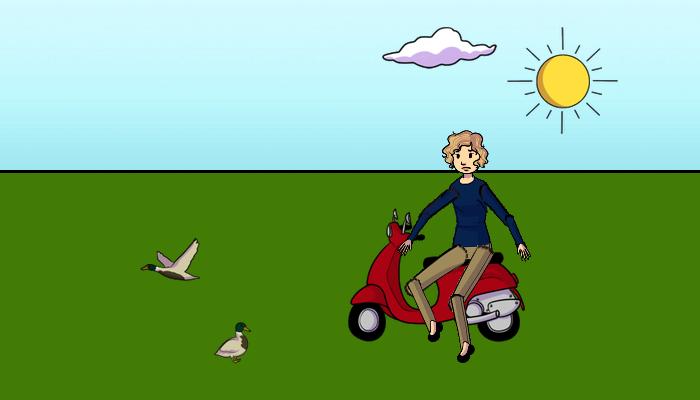}\\
    \includegraphics[width=42mm]{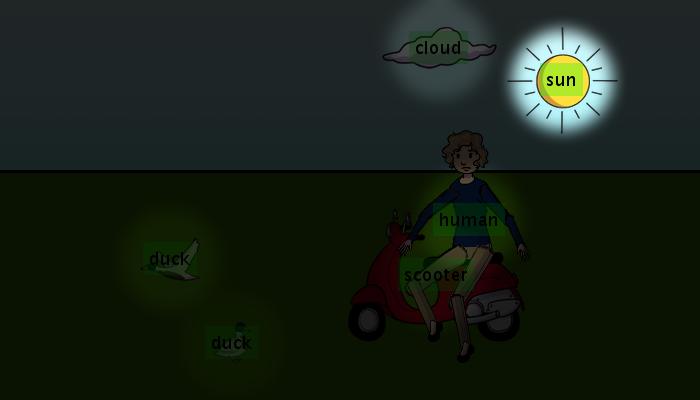}\\
  }&
    \begin{minipage}{\linewidth}\footnotesize\input{"7166-837.txt"}\end{minipage}
    \vspace{7mm}
    \hangBox{\includegraphics[width=40mm]{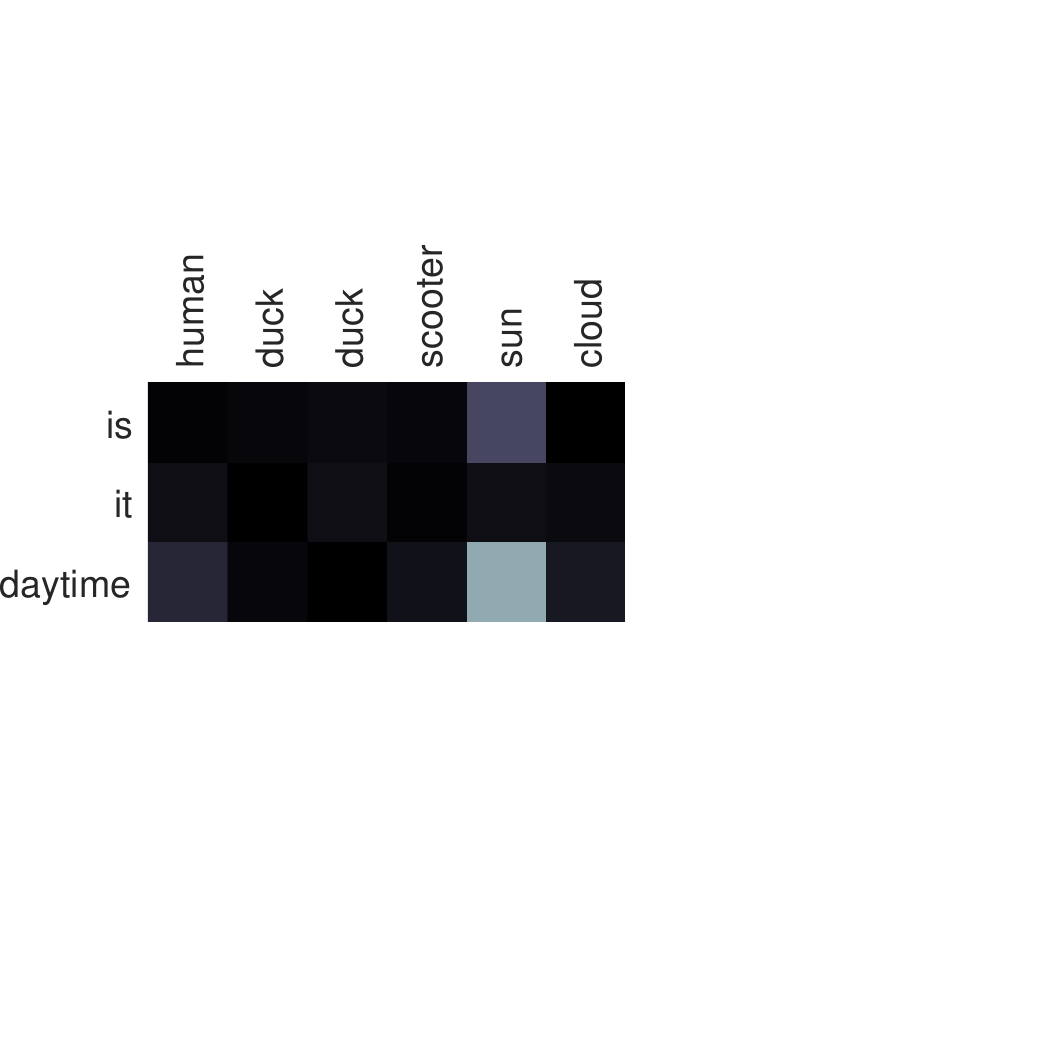}}\vspace{-7mm}
  &
  \vspace{-12mm}
  \hangBox{
    \includegraphics[width=42mm]{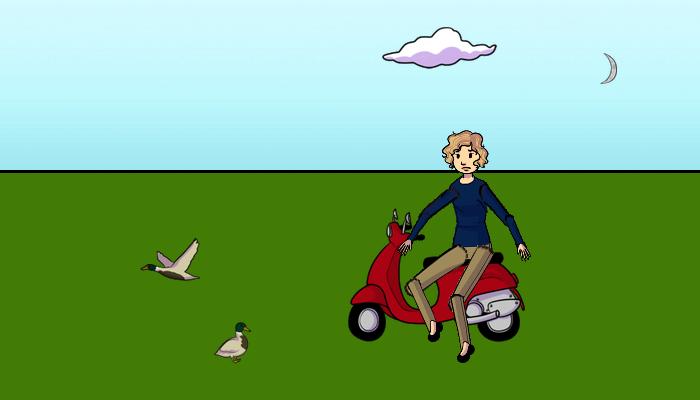}\\
    \includegraphics[width=42mm]{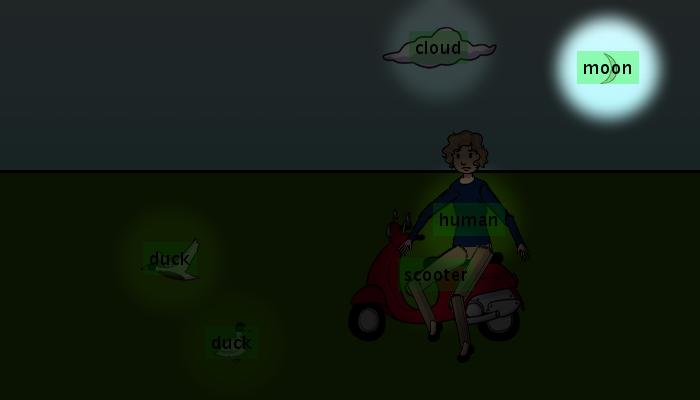}\\
  }&
    \begin{minipage}{\linewidth}\footnotesize\input{"7167-900008371.txt"}\end{minipage}
    \vspace{7mm}
    \hangBox{\includegraphics[width=40mm]{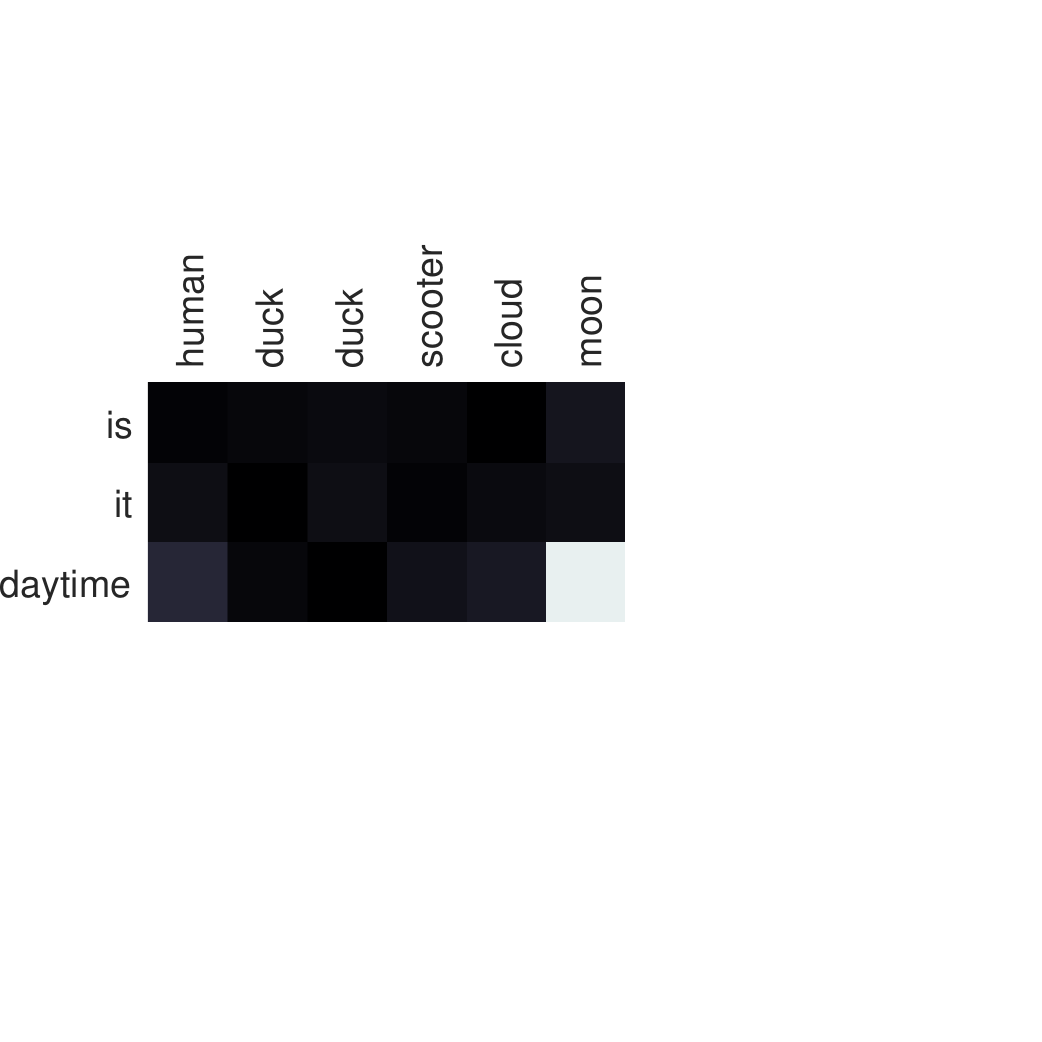}}\vspace{-7mm}

  \end{tabularx}
  \vspace{-9pt}
  \caption{Qualitative results on the ``abstract scenes'' dataset (top row) and on ``balanced'' pairs (middle and bottom row). We show the input scene, the question, the predicted answer, and the correct answer when the prediction is erroneous. We also visualize the matrices of matching weights (\eq\ref{eq:weights}, brighter correspond to higher values) between question words (vertically) and scene objects (horizontally). The matching weights are also visualized over objects in the scene, after summation over words, giving an indication of their estimated relevance. The ground truth object labels are for reference only, and not used for training or inference.}
  \label{fig:qualitative}
  \vspace{-9pt}
\end{figure*}

\section{Conclusions}

We presented a deep neural network for visual question answering that processes graph-structured representations of scenes and questions. This enables leveraging existing natural language processing tools, in particular pretrained word embeddings and syntactic parsing. The latter showed significant advantage over a traditional sequential processing of the questions, \eg with LSTMs. In our opinion, VQA systems are unlikely to learn everything from question/answer examples alone. We believe that any significant improvement in performance will require additional sources of information and supervision. Our explicit processing of the language part is a small step in that direction. It has clearly shown to improve generalization without resting entirely on VQA-specific annotations. We have so far applied our method to datasets of clip art scenes. Its direct extension to real images will be addressed in future work, by replacing nodes in the input scene graph with proposals from pretrained object detectors.

\newpage

{\small\bibliographystyle{ieee}\bibliography{Bibliography}}
\clearpage

\appendix

\section*{Supplementary material}

\section{Implementation}
\label{implDetails}

\noindent
We provide below practical details of our implementation of the proposed method.

\begin{enumerate}[topsep=0pt,itemsep=-1ex,partopsep=1ex,parsep=1.5ex,label={\tiny$\bullet$},leftmargin=2.0ex]
\item Size of vector embeddings of node features, edge features, and all hidden states within the network: $H$\sheq300. Note that smaller values such as $H$\sheq200 also give very good results (not reported in this paper) at a fraction of the training time.
\item Number of recurrent iterations to update graph node representations: $T^\symbQ$=$T^\symbS$\sheq4. Anecdotally, we observed that processing the scene graph benefits from more iterations than the question graph, for which performance nearly saturates with 2 or more iterations. As reported in the ablative evaluation (\tab\ref{tab:balanced}), the use of at least a single iteration has a stronger influence than its exact number.
\item All weights except word embeddings are initialized randomly following \cite{glorot2010understanding}.
\item Word embeddings are initialized with Glove vectors \cite{pennington2014glove} of dimension 300 available publicly \cite{gloveWebsite}, trained for 6 billion words on Wikipedia and Gigaword. The word embeddings are fine-tuned with a learning rate of $1/10$ of the other weights.
\item Dropout with ratio 0.3 is applied between the weighted sum over scene elements (\eq\ref{eq:out2}) and the final classifier (\eq\ref{eq:out3}).
\item Weights are optimized with Adadelta \cite{zeiler2012adadelta} with mini-batches of 128 questions. We run optimization until convergence (typically 20 epochs on the ``abstract scenes'', 100 epochs on the ``balanced'' dataset) and report performance on the test set from the epoch with the highest performance on the validation set (measured by VQA score on the ``abstract scenes'' dataset, and accuracy over pairs on the ``balanced'' dataset).

\item The edges between word nodes in the input question graph are labeled with the dependency labels identified by the Stanford parser \cite{demarneffe2008parser,stanfordParserWebsite}. These dependencies are directed, and we supplement all of them with their symmetric, albeit tagged with a different set of labels. The output of the parser includes the propagation of conjunct dependencies (its default setting). This yields quite densely connected graphs.

\item The input features of the object nodes are those directly available in the datasets. They represent: the object category (human, animal, small or large object) as one one-hot vector, the object type (table, sun, dog window, ...) as a one-hot vector, the expression/pose/type (various depictions being possible for each object type) as a one-hot vector, and 10 scalar values describing the pose of human figures (the X/Y position of arms, legs, and head relative to the torso). They form altogether a feature vector of dimension 159. The edge features between objects represent: the signed difference in their X/Y position, the inverse of their absolute difference in X/Y position, and their relative position on depth planes as +1 if closer (potentially occluding the other), -1 otherwise. 

\item All input features are normalized for zero mean and unit variance.
\item When training for the ``balanced'' dataset, care is taken to keep each pair of complementary scenes in a same mini-batch when shuffling training instances. This has a noticeable effect on the stability of the optimization.
\item In the open-ended setting, the output space is made of all answers that appear at least 5 times in the training set. These correspond to 623 possible answers, which cover 96\% of the training questions.
\item Our model was implemented in Matlab from scratch. Training takes in the order of 5 to 10 hours on one CPU, depending on the dataset and on the size $H$ of the internal representations.


\end{enumerate}

\section{Additional details}

\noindent\textbf{\textit{Why do we choose to focus on abstract scenes ? Does this method extend to real images ?}}



\noindent
The balanced dataset of abstract scenes was the only one allowing evaluation free from dataset biases. Abstract scenes also enabled removing confounding factors (visual recognition). It is not unreasonable to view the scene descriptions (provided with abstract scenes) as the output of a ``perfect'' vision system. The proposel model could be extended to real images by building graphs of the images where scene nodes are candidates from an object detection algorithm.

\vspace{5pt}
\noindent\textbf{\textit{The multiple-choice (M.C.) setting should be easier than open-ended (O.E.). Therefore, why is the accuracy not better for binary and number questions in the M.C setting (rather than O.E.) ?}}

\noindent
This intuition is incorrect in practice. The wording of binary and number questions (``\textit{How many ...}'') can easily narrow down the set of possible answers, whether evaluated in a M.C. or O.E. setting. One thus cannot qualify one as strictly easier than the other. Other factors can then influence the performance either way. Note also that, for example that most choices of number questions are not numbers.

\vspace{5pt}
\noindent\textbf{\textit{In Table 1, why is there a large improvement of the metric over balanced pairs of scenes, but not of the metric over individual scenes ?}}

\noindent
The metric over pairs is much harder to satisfy and should be regarded as more meaningful. The other metric (over scenes) essentially saturates at the same point between the two methods.

\vspace{5pt}
\noindent\textbf{\textit{How are precison/recall curves helping better understand model compared to a simple accuracy number ?}}

\noindent
A P/R curve shows the confidence of the model in its answers. A practical VQA system will need to provide an indication of certainty, including the possibility of ``I don't know''. Reporting P/R is a step in that direction. P/R curves also contain more information and can show differences between methods (\eg Fig.3 left) that may otherwise not be appreciable through an aggregate metric.

\vspace{5pt}
\noindent\textbf{\textit{Why is attention computed with pre-GRU node features ?}}

\noindent
This performed slightly better than the alternative. The intuition is that the identity of each node is sufficient, and the context (transfered by the GRU from neighbouring nodes) is probably less useful to compute attention.

\vspace{5pt}
\noindent\textbf{\textit{Why are the largest performance gains obtained with ``number'' questions ?}}

\noindent
We could not draw definitive conclusions. Competing methods seem to rely on dataset biases (predominance of \textit{2} and \textit{3} as answers). Ours was developed (cross-validated) for the balanced dataset, which requires \emph{not} to rely on such biases, and may simply be better at utilizing the input and not biases. This may in turn explain minimal gains on other questions, which could benefit from using biases (because of a larger pool of reasonable answers).

\section{Additional results}

We provide below additional example results in the same format as in \fig\ref{fig:qualitative}.

\clearpage

\subsection{Additional results: abstract scenes dataset}
\vspace{10mm}


\resultsAttentionAbstractPageSix{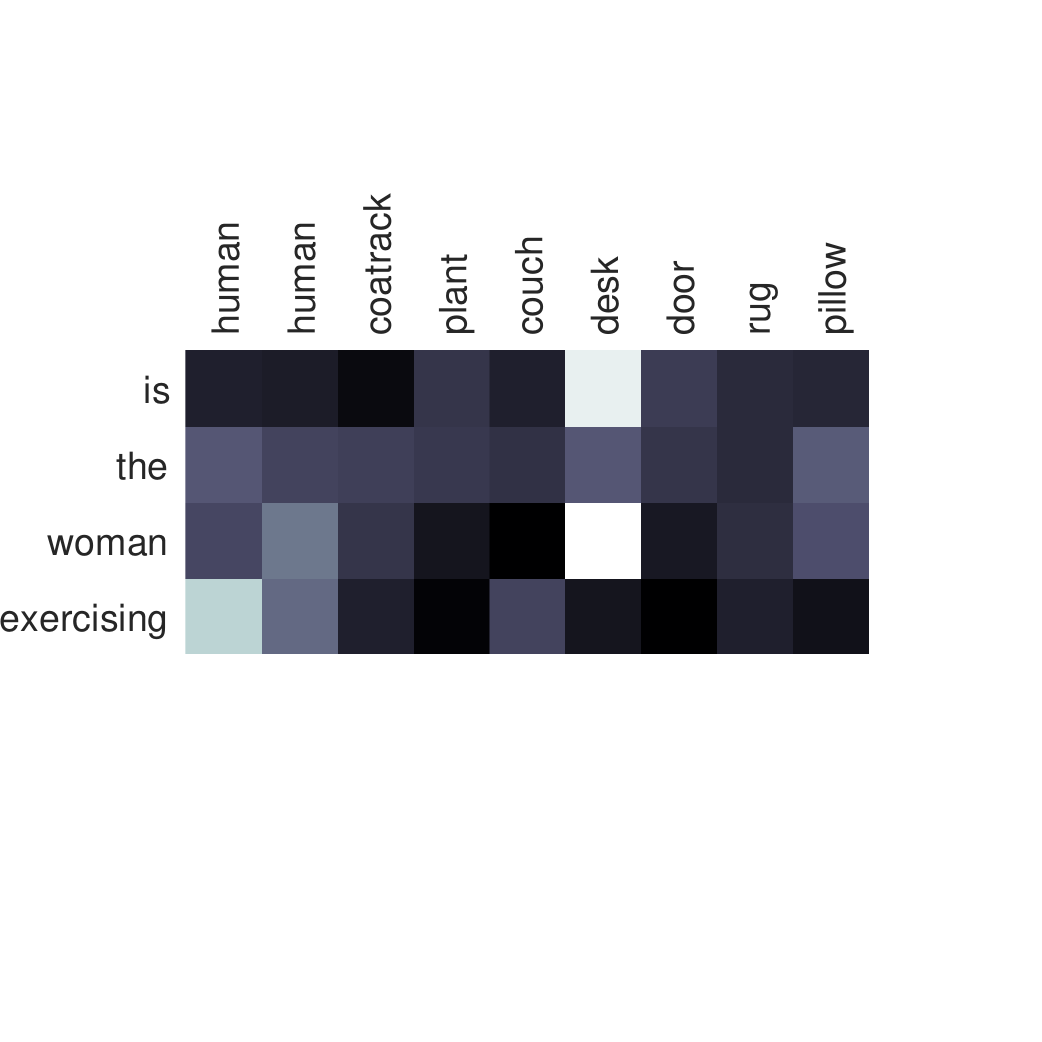}{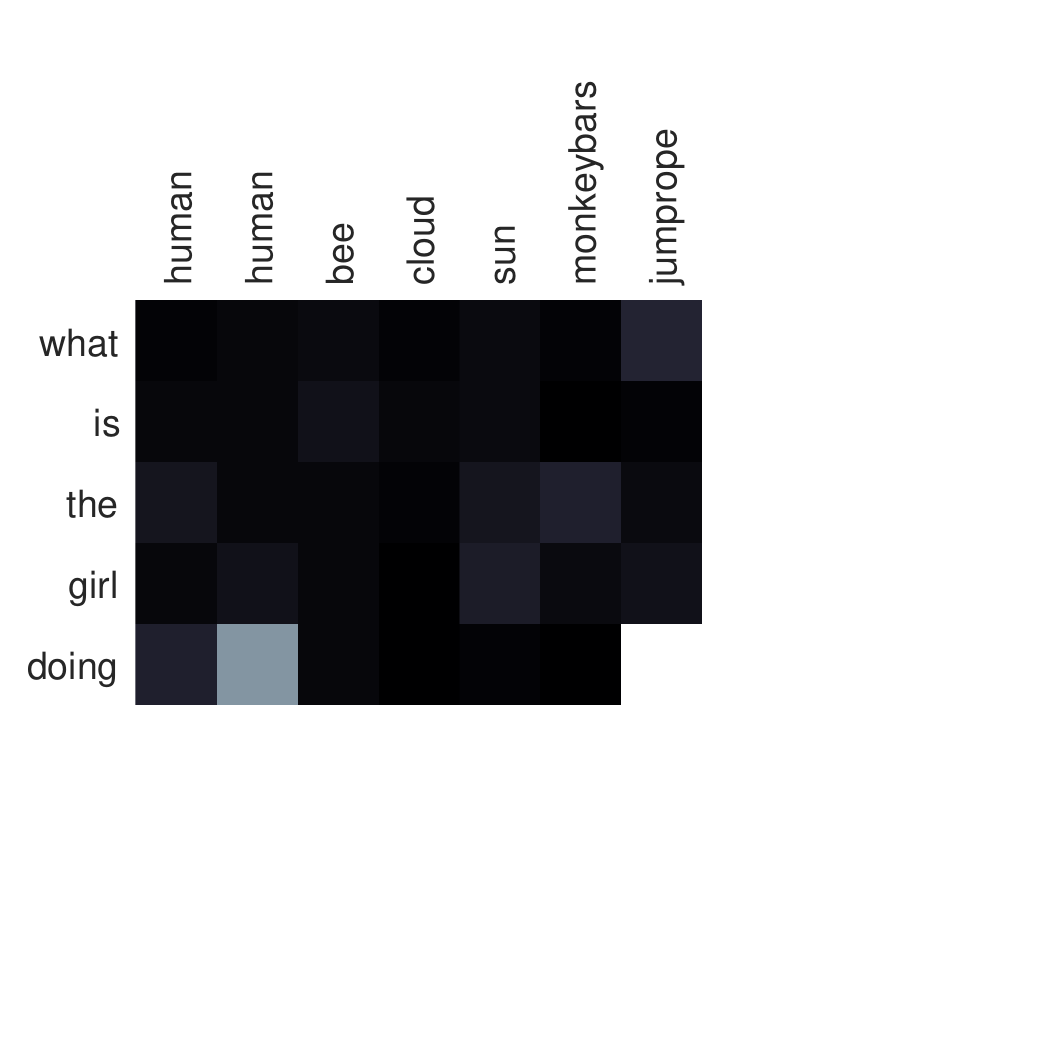}{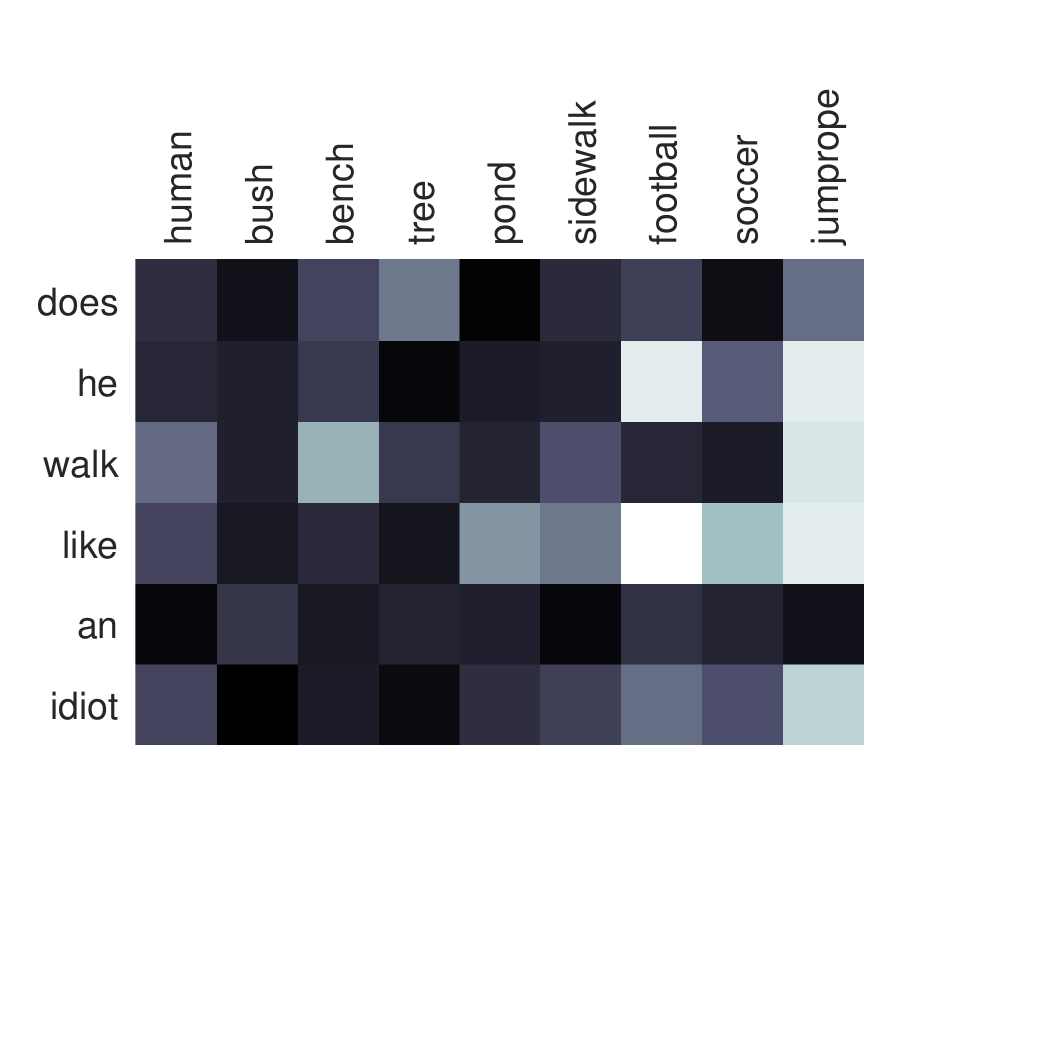}{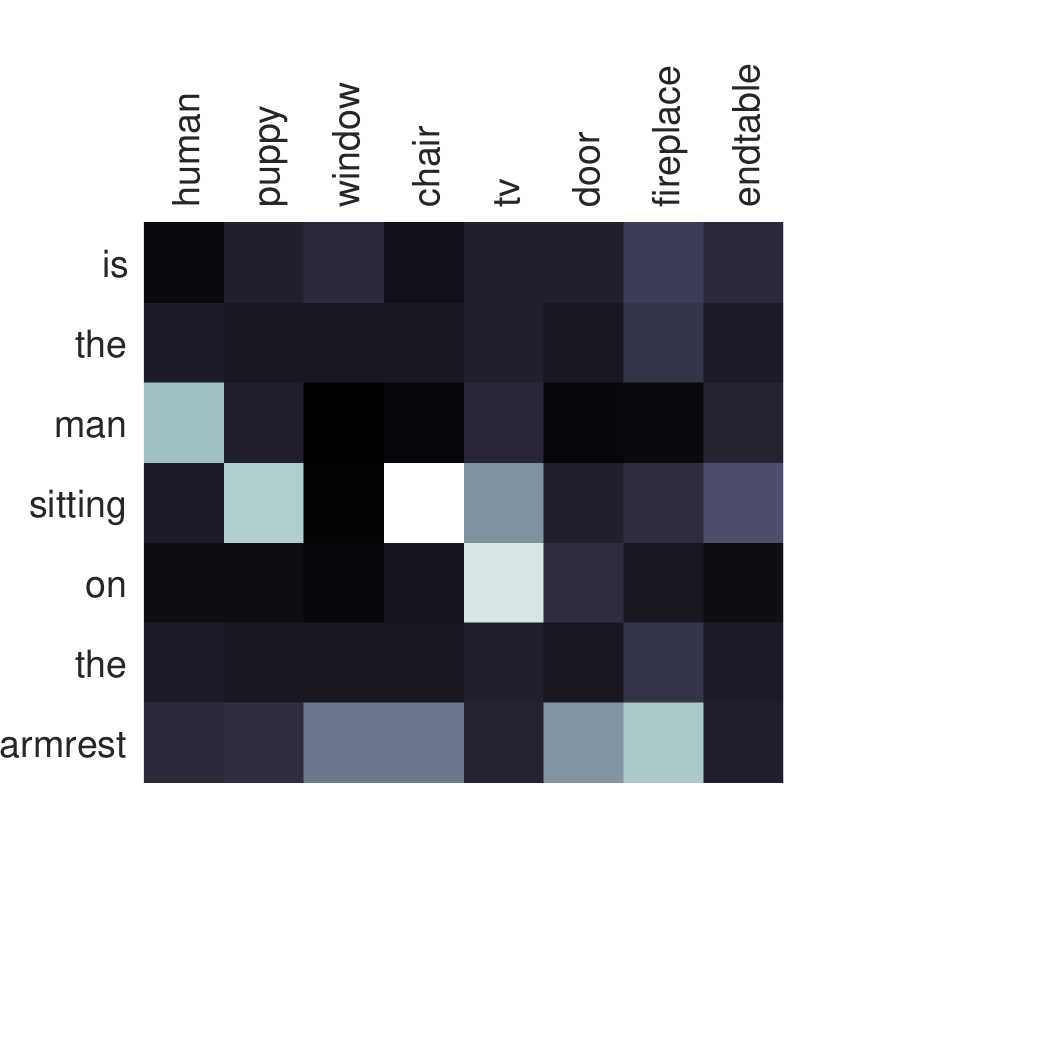}{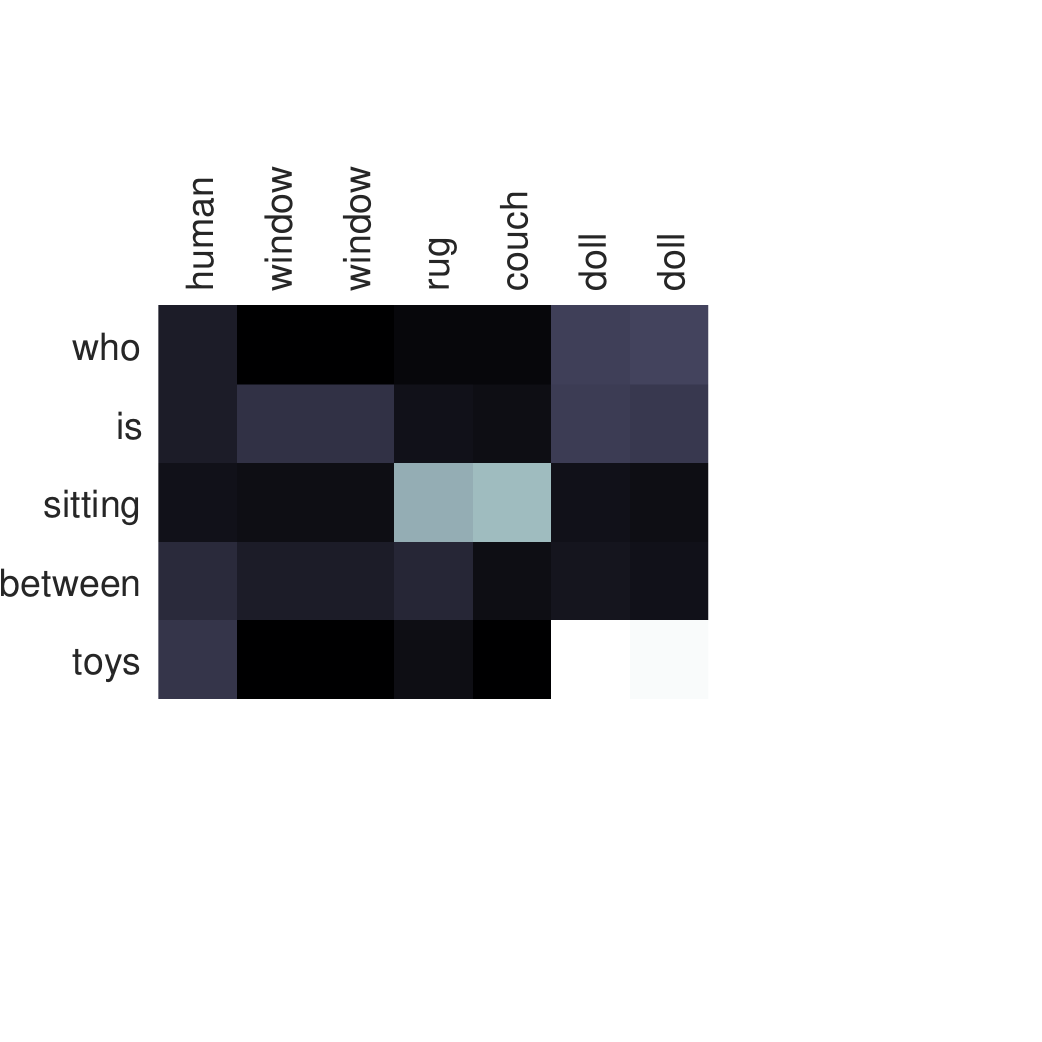}{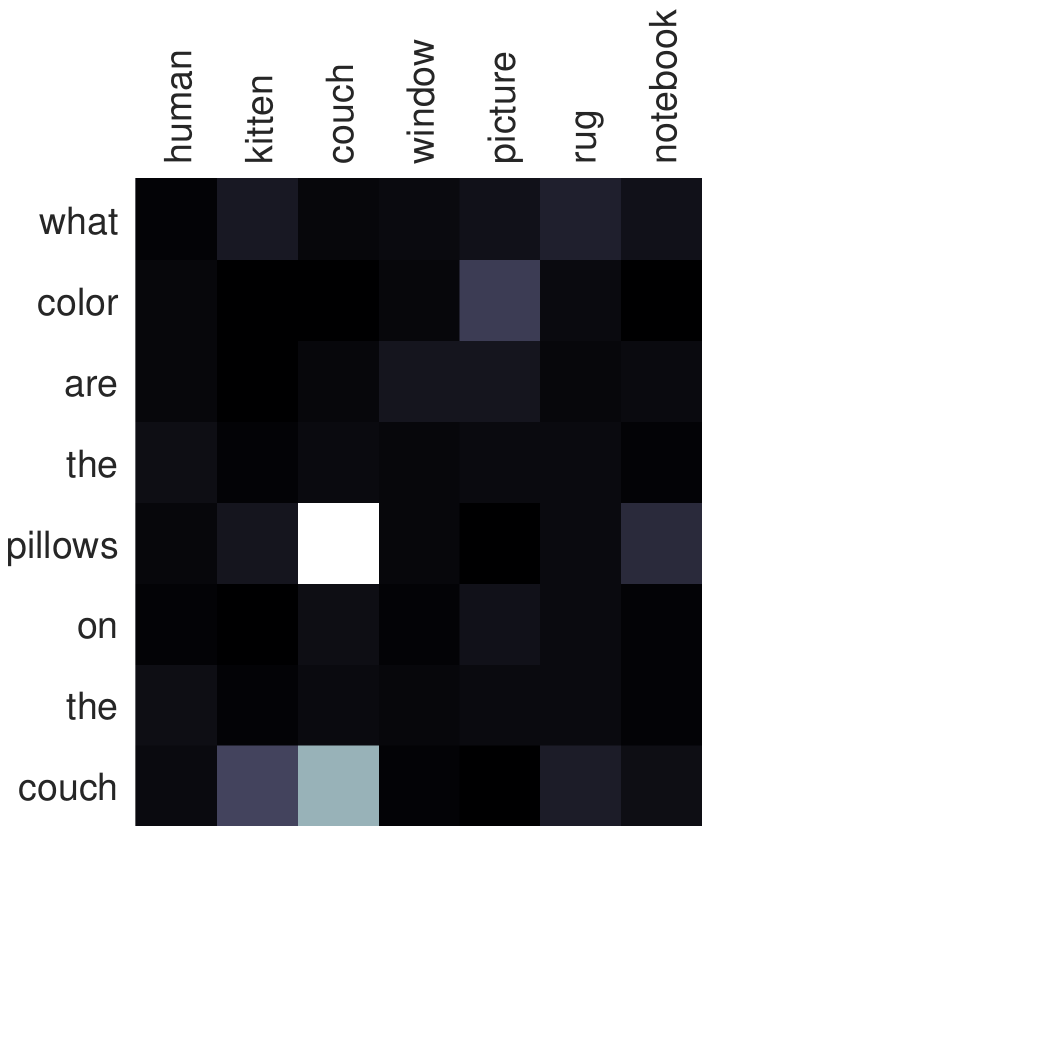} \clearpage 
\resultsAttentionAbstractPage{29551}{29627}{29685}{29758}{29804}{29819}{29865}{29907} \clearpage
\resultsAttentionAbstractPage{22937}{23128}{23273}{23301}{23328}{23359}{23429}{23521} \clearpage

\clearpage
\subsection{Additional results: balanced dataset}
\vspace{10mm}

\resultsAttentionBalancedPageSix{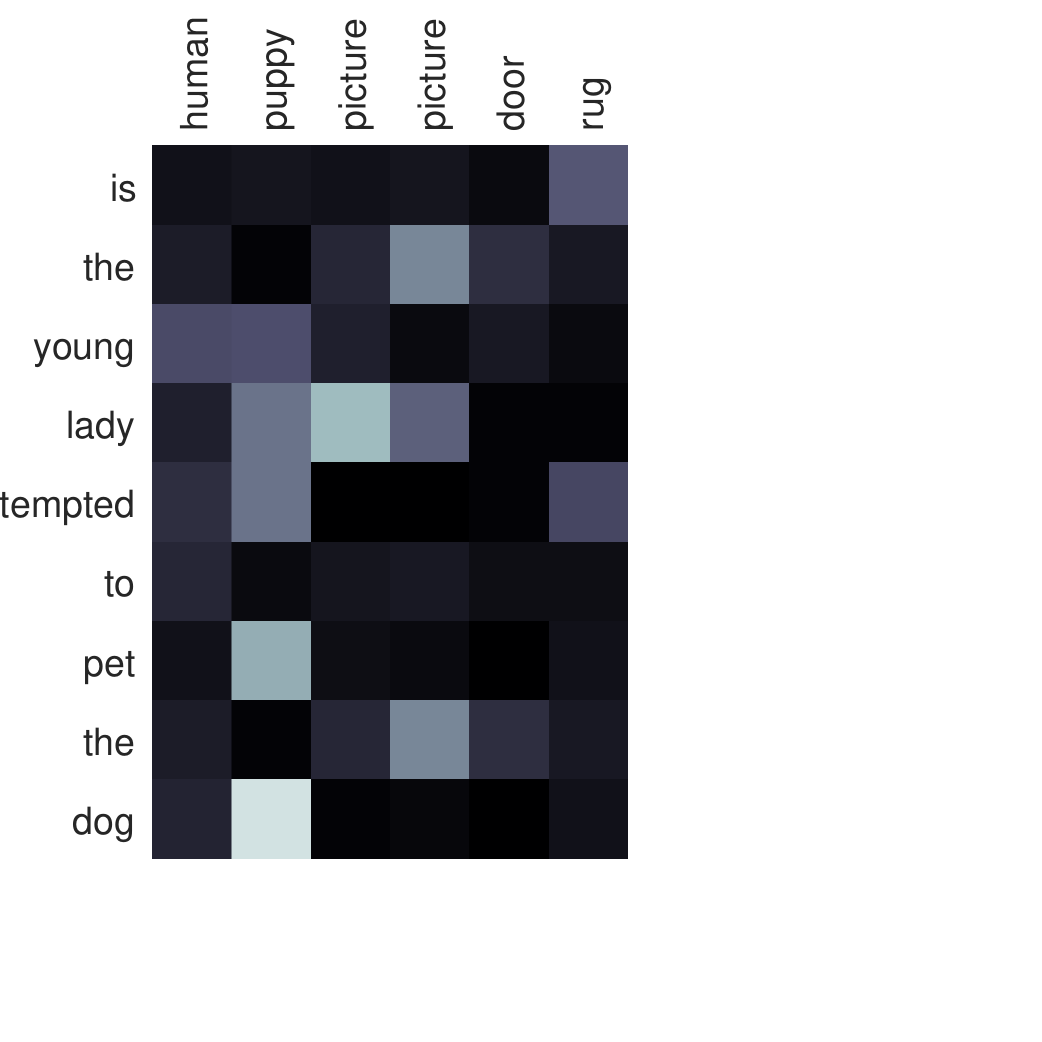}{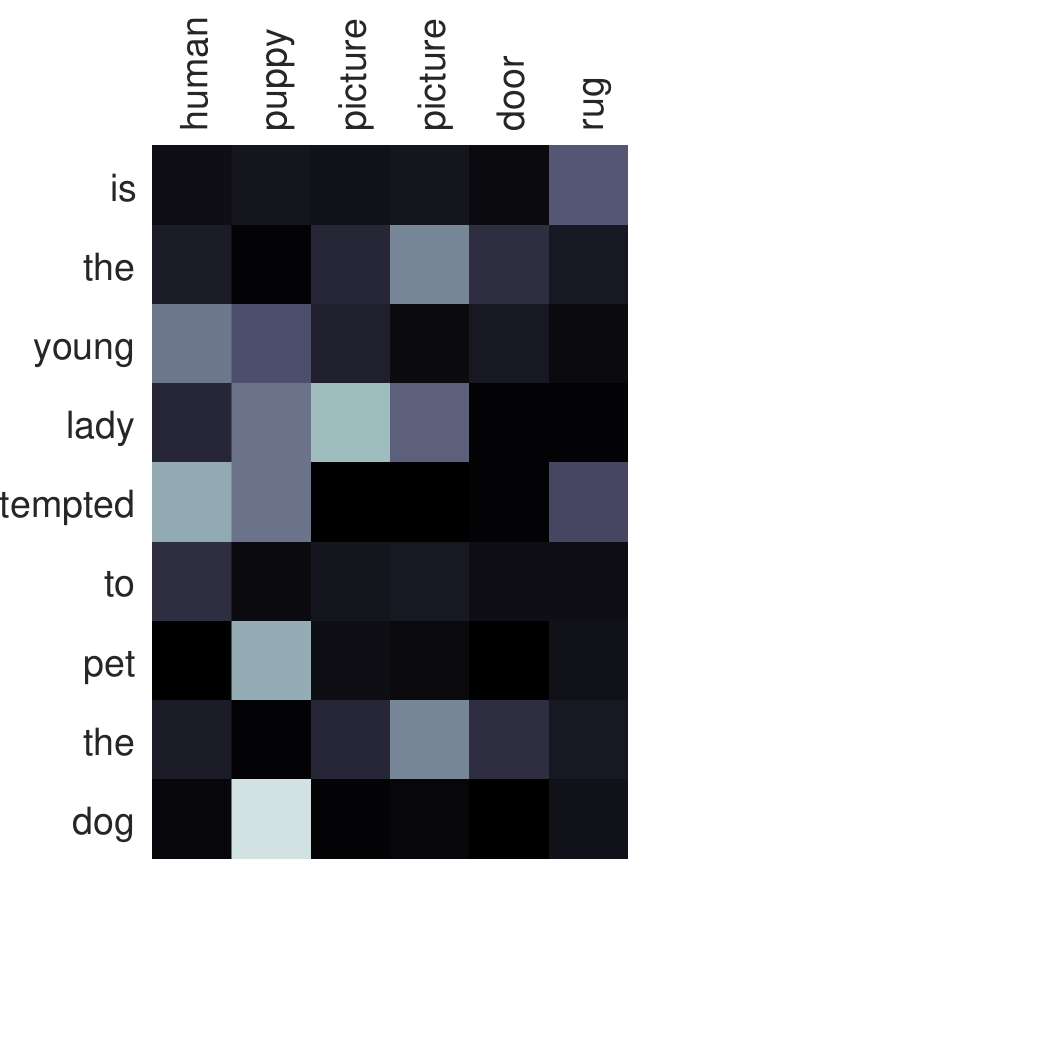}{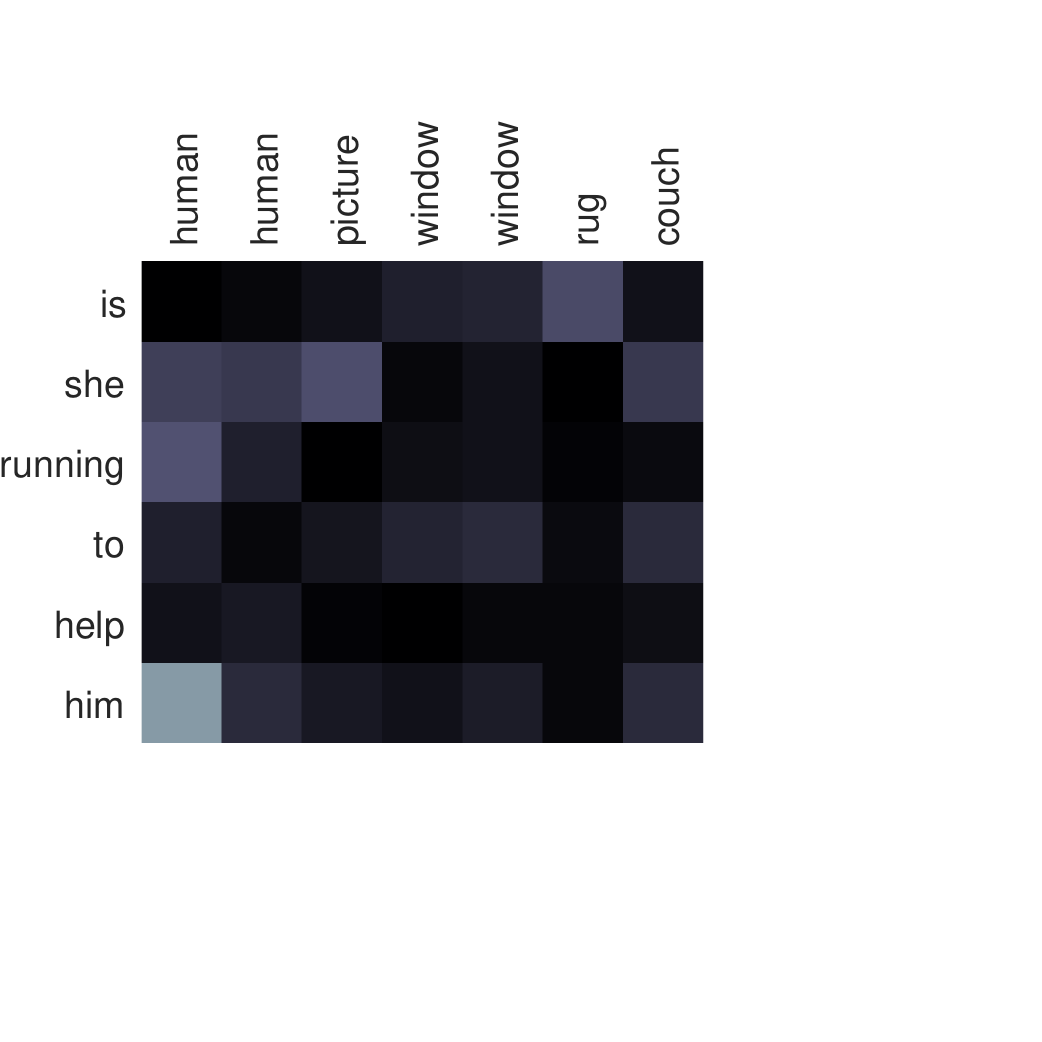}{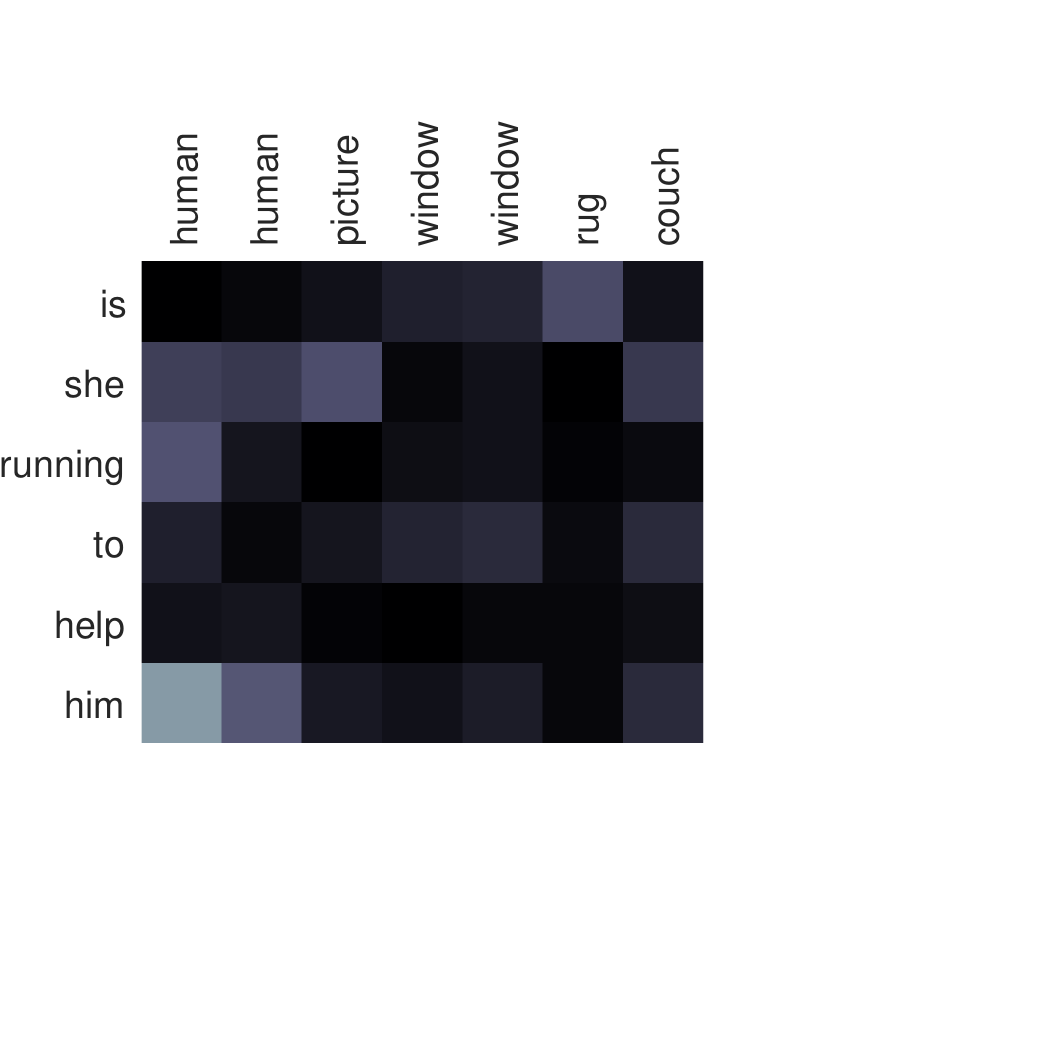}{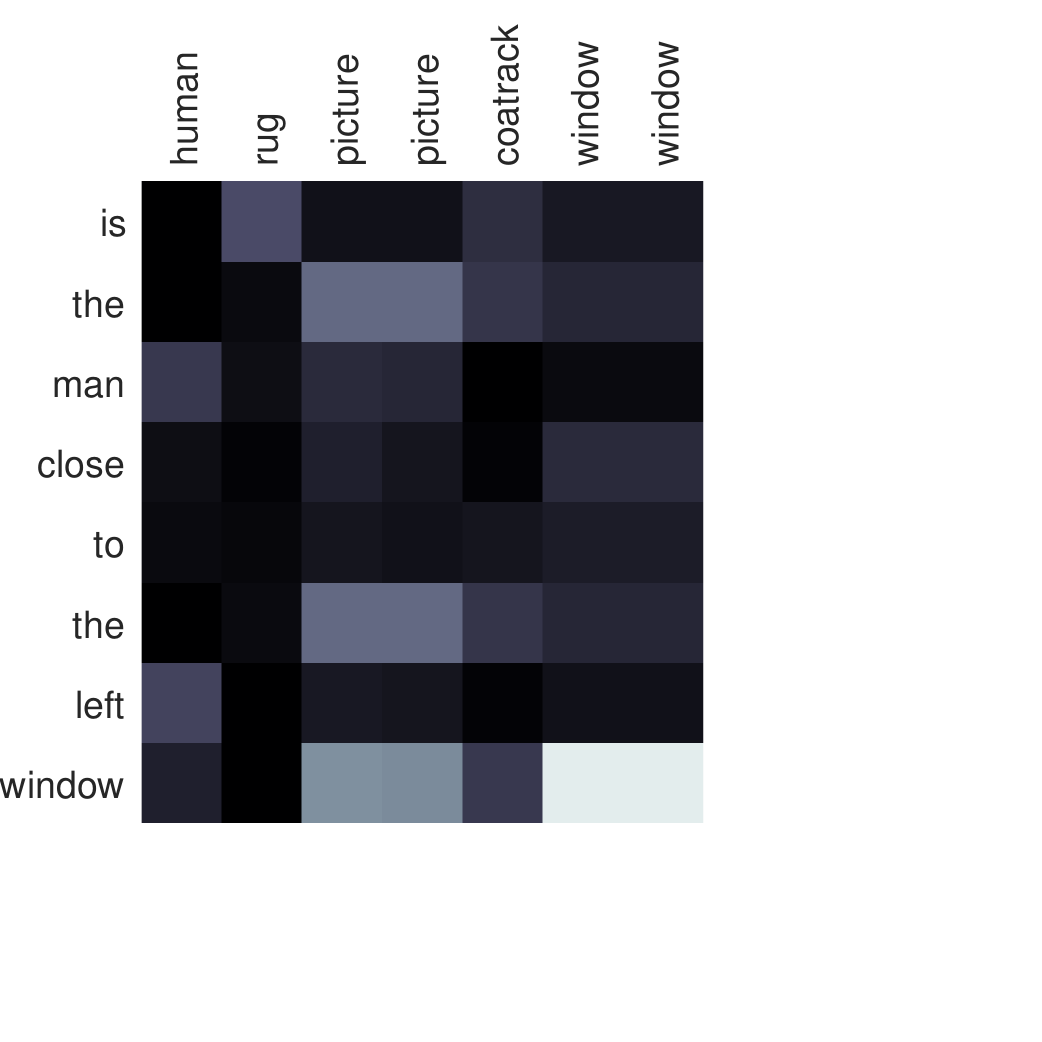}{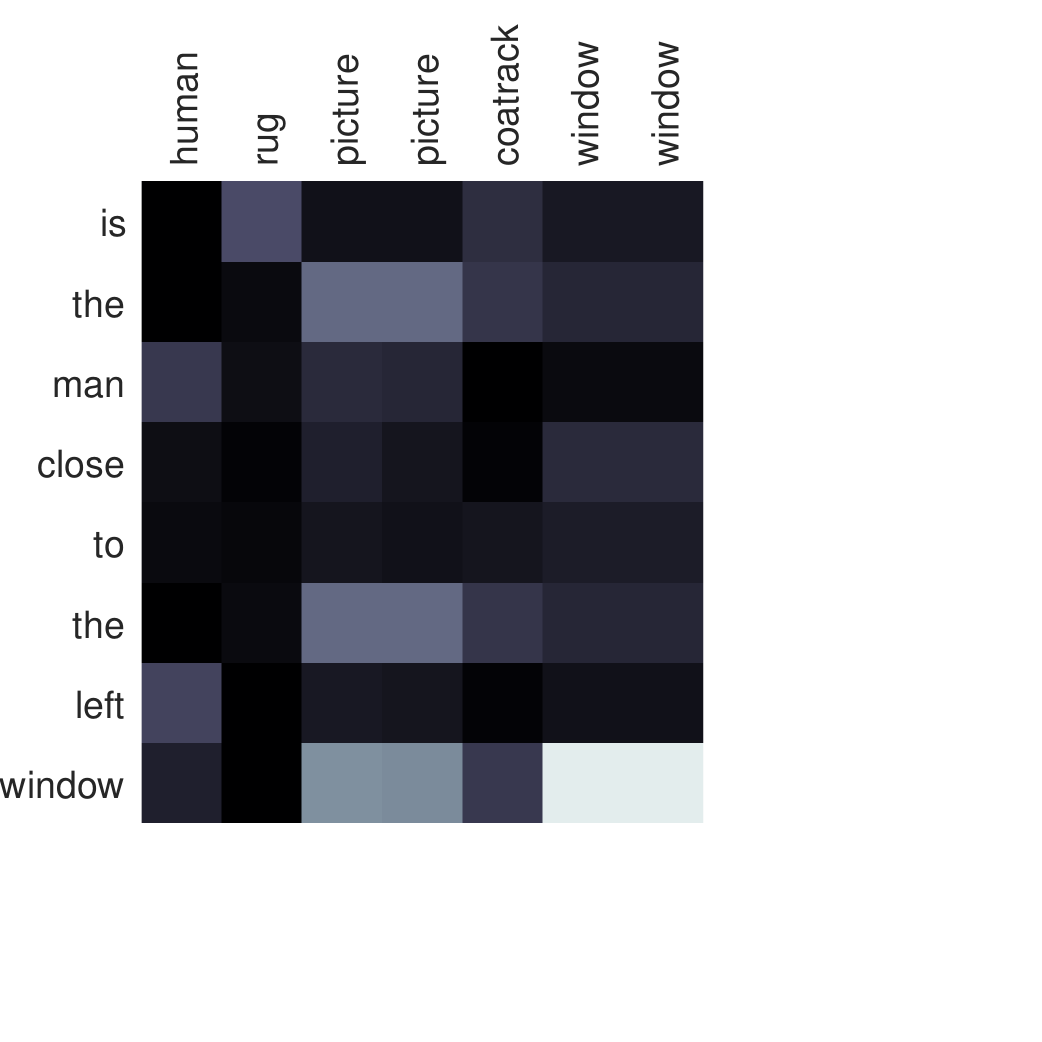} \clearpage 
\resultsAttentionBalancedPage{8057-11746}{8058-900117462}{8458-5536}{8459-900055360}{8466-14578}{8467-900145780}{9149-3819}{9150-900038192} \clearpage
\resultsAttentionBalancedPage{14965-5372}{14966-900053720}{14967-10677}{14968-900106771}{15752-7867}{15753-900078671}{16088-4845}{16089-900048451}

\end{document}